\documentclass{article}

\usepackage{microtype}
\usepackage{graphicx}
\usepackage{subfigure}
\usepackage{booktabs} 

\usepackage{hyperref}

\usepackage[accepted]{arxiv0301}

\usepackage{amsmath}
\usepackage{amssymb}
\usepackage{mathtools}
\usepackage{amsthm}

\usepackage[capitalize,noabbrev]{cleveref}

\usepackage{longtable}

\theoremstyle{plain}

\theoremstyle{definition}

\theoremstyle{remark}

\usepackage[textsize=tiny]{todonotes}

\icmltitlerunning{InfiAgent-DABench: Evaluating Agents on Data Analysis Tasks}

\begin{document}

\twocolumn[
\icmltitle{\raisebox{-.2\height}{\includegraphics[width=0.7cm]{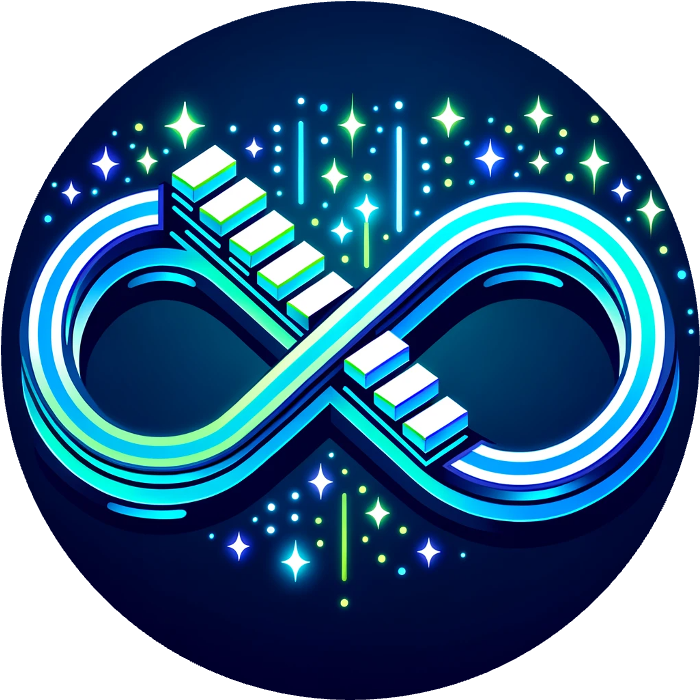}} InfiAgent-DABench: Evaluating Agents on Data Analysis Tasks}



\icmlsetsymbol{equal}{*}

\begin{icmlauthorlist}
\icmlauthor{Xueyu Hu}{zju}
\icmlauthor{Ziyu Zhao}{zju}
\icmlauthor{Shuang Wei}{rit}
\icmlauthor{Ziwei Chai}{zju}
\icmlauthor{Qianli Ma}{bytedance}
\\
\icmlauthor{Guoyin Wang}{bytedance}
\icmlauthor{Xuwu Wang}{bytedance}
\icmlauthor{Jing Su}{bytedance}
\icmlauthor{Jingjing Xu}{bytedance}
\icmlauthor{Ming Zhu}{bytedance}
\icmlauthor{Yao Cheng}{bytedance}
\icmlauthor{Jianbo Yuan}{bytedance}
\\
\icmlauthor{Jiwei Li}{zju}
\icmlauthor{Kun Kuang}{zju}
\icmlauthor{Yang Yang}{zju}
\icmlauthor{Hongxia Yang}{bytedance}
\icmlauthor{Fei Wu}{zju}
\end{icmlauthorlist}

\icmlaffiliation{zju}{Zhejiang University}
\icmlaffiliation{rit}{Rochester Institute of Technology}
\icmlaffiliation{bytedance}{ByteDance Inc.}

\icmlcorrespondingauthor{Jingjing Xu}{gracexu@bytedance.com}

\icmlkeywords{Machine Learning, ICML}

\vskip 0.3in
]



\printAffiliationsAndNotice{}  

\begin{abstract}
In this paper, we introduce \textbf{InfiAgent-DABench}, the first benchmark specifically designed to evaluate LLM-based agents on data analysis tasks. 
These tasks require agents to end-to-end solving complex tasks by interacting with an execution environment.
This benchmark contains DAEval, a dataset consisting of 257 data analysis questions derived from 52 CSV files, and an agent framework which incorporates LLMs to serve as data analysis agents for both serving and evaluation.
Since data analysis questions are often open-ended and hard to evaluate without human supervision, we adopt a format-prompting technique to convert each question into a closed-form format so that they can be automatically evaluated. Our extensive benchmarking of 34 LLMs uncovers the current challenges encountered in data analysis tasks. 
In addition, building on top of our agent framework, we develop a specialized agent, DAAgent, which surpasses GPT-3.5 by 3.9\% on DABench.
Evaluation datasets and toolkits for InfiAgent-DABench are released at \url{https://github.com/InfiAgent/InfiAgent}.
\end{abstract}
\section{Introduction}

Large language model-based agents (LLM-based agents) has been one of the most popular concepts recently in AI society \citep{liu2023wants, wang2023survey, zhao2023depth, xi2023rise}, with the belief that these agents are a prototype of Artificial General Intelligence (AGI). These agents can reason \citep{huang2023reasoning, wei2022chain, kojima2022large}, plan \citep{xu2023rewoo, sun2023adaplanner, yao2022react}, memorize \citep{liang2023unleashing, zhang2023memory} and even use tools \citep{qin2023toolllm, qin2023tool}, by leveraging the ability of large language models (LLMs). Nowadays, a large number of LLM-based agents have been developed, \emph{e.g.}, AutoGPT \citep{2023autogpt}, BabyAGI \citep{2023babyagi} and AgentGPT \citep{2023agentgpt}.

\begin{figure*}[t]
  \centering
  \includegraphics[width=0.95\textwidth]{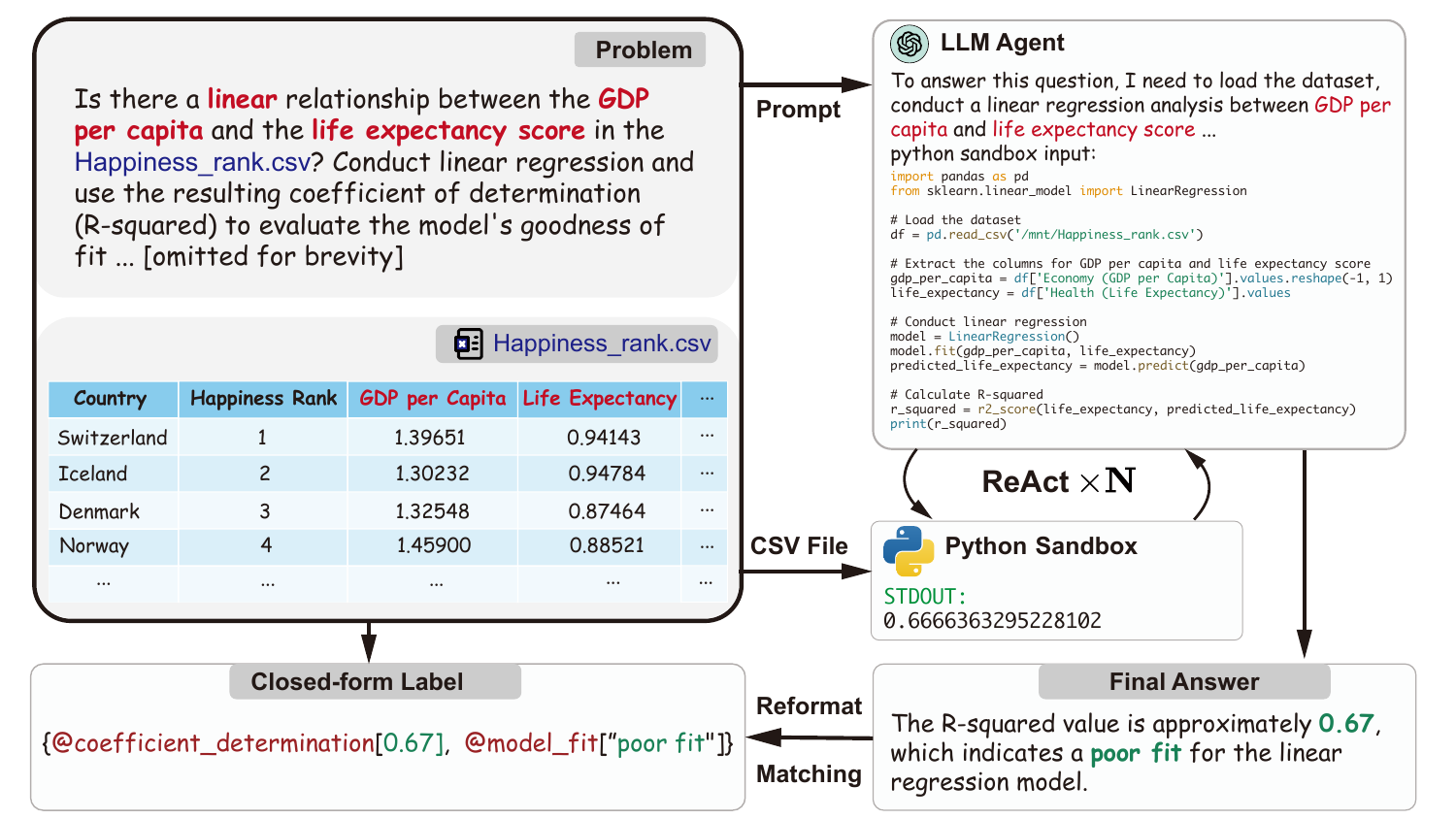}
  \caption{\textbf{An illustration of evaluation on InfiAgent-DABench.} A LLM-based agent is prompted with a data analysis question and corresponding CSV file. It plans, writes code, calls a Python sandbox and concludes to solve the problem in a ReAct way. The final answer is reformatted into specific format and matched with the closed-form label.}
  \label{fig:framework}
  \vspace{-10pt}
\end{figure*}

Specifically, data analysis tasks have been found to be challenging but practically useful problems for LLM-based agents \citep{lai2023ds, xie2023openagents, bai2023qwen}.
Data analysis is a systematic process of examining, cleaning, transforming, and modeling data to discover useful information, inform conclusions, and support decision-making \citep{hardy2004handbook}. 
It is applied across diverse domains including business intelligence \citep{covington2016analytics}, healthcare \citep{reddy2015healthcare}, finance \citep{koop2022analysis}, and scientific research \citep{sabina2020scientific}. 
A typical data analysis problem requires the solver to interact with data files and the code executor to obtain a solution, which requires the harmonization ability of LLMs in natural language and code, and is suitable for LLM-based agents to solve \citep{yang2024if}. 
Figure \ref{fig:framework} shows how a LLM-based agent solve a problem with a CSV file as input, following the ReAct \citep{yao2022react} approach to plan, write code, call external Python sandbox to execute and concluding the output from the sandbox to solve the problem. 
OpenAI's Advanced Data Analysis\footnote{\url{https://openai.com/blog/chatgpt-plugins\#code-interpreter}} (ADA) is the first commercial LLM-based agent that effectively helps users analyze data files with the execution of programming code.
On top of OpenAI ADA, many developers have also formulated open-source agents for data analysis, including Open Interpreter \citep{2023openinterpreter}, Open Agents \citep{xie2023openagents}, Qwen-Agent \citep{bai2023qwen} and Taskweaver \citep{qiao2023taskweaver}. 

However, a comprehensive benchmark for evaluating agents on data analysis is still missing in the community, which motivates us to build InfiAgent-DABench\footnote{The series of works is named as InfiAgent, which is included as part of the comprehensive name.}, to our knowledge, the first benchmark for evaluating agents on data analysis tasks. Although there are many benchmarks for code-completion tasks, \emph{e.g.}, HumanEval \citep{chen2021evaluating}, MBPP \citep{austin2021program}, and DS-1000 \citep{lai2023ds}, most of them are not suitable for evaluating LLM-based agents which require complex abilities, such as planning, self-debugging. For instance, DS-1000 only requires models to insert or complete a small part of code snippet. In InfiAgent-DABench, we aim to evaluate end-to-end LLM-based agents covering abilities like task planning, writing code, leveraging Python sandbox, self-debugging 
\citep{chen2023teaching} if needed, and concluding to solve problems.

\begin{figure*}[t]
  \centering
  \includegraphics[width=\textwidth]{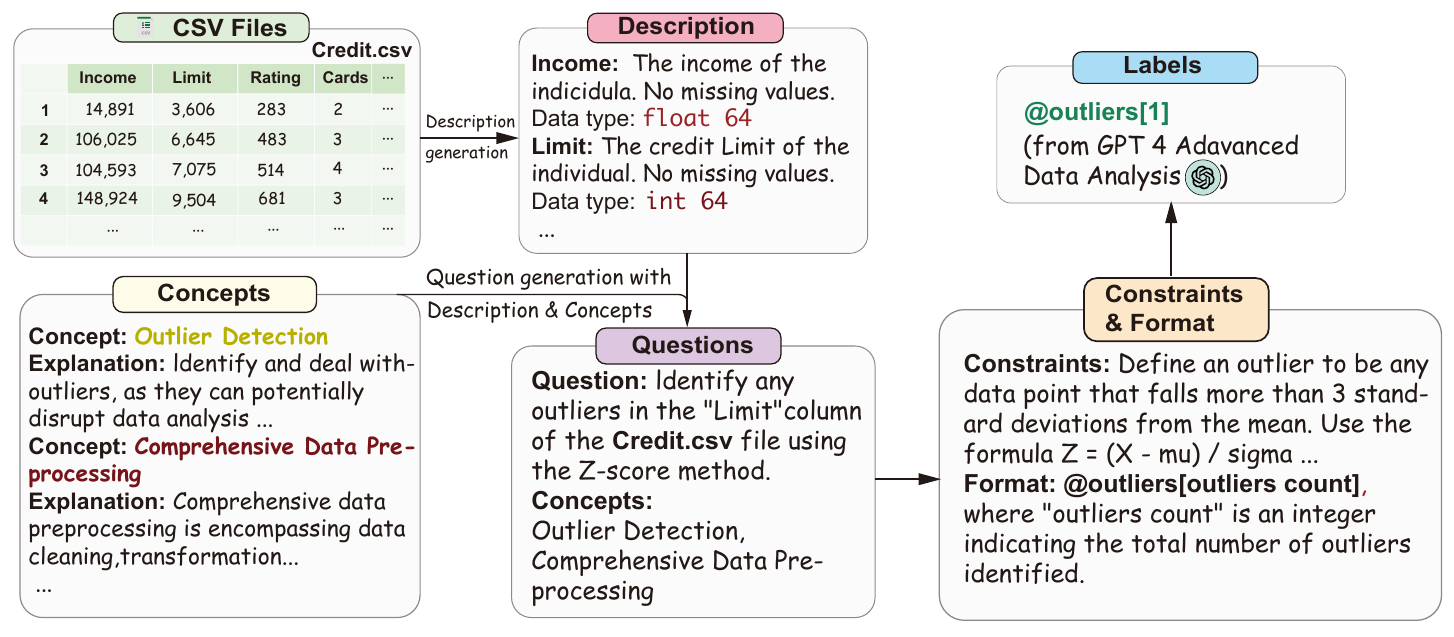}
  \caption{\textbf{The workflow of DAEval construction.} Data analysis questions are generated with GPT-4 based on the description of CSV files and pre-defined key concepts. Then, we generate constraints and format requirements to make questions closed-form for precise evaluation. The labels are obtained from OpenAI ADA. All these are undergone strict human assessment and filtering.}
  \label{fig:workflow dataset construction}
  \vspace{-10pt}
\end{figure*}

In this paper, we build InfiAgent-DABench for data analysis tasks, which includes 257 questions associated with 52 CSV files, covering realistic data analysis demands and a wide range of domains. We crawl CSV files from GitHub and instruct GPT-4 to generate open-ended questions based on the file and several key concepts for data analysis obtained from expert interviews. Then, as shown in Figure \ref{fig:workflow dataset construction}, constraints and format requirements containing detailed strict on methods and series of ``@answer\_name[answer]'' as a special format the final answer must follow are generated through a format-prompting technique. Models can be automatically evaluated in this way as the answer is unique and can be easily parsed and matched. 

The dataset is strictly assessed by human experts in multiple dimensions and all unqualified samples are filtered. In addition, to facilitate evaluation, we construct an agent framework to support LLMs to perform data analysis tasks. We evaluated 34 state-of-the-art LLMs and found they are confronted with challenges on handling data analysis tasks. 

Moreover, to facilitate the construction of open-source LLMs with data analysis agent capability, we develop an automated method to construct an instruction-tuning dataset DAInstruct for data analysis tasks. Sharing the same merits as tool-learning instruction datasets \citep{qin2023toolllm, patil2023gorilla, li2023api} for real-world APIs, we first construct instructions based on data analysis keywords and real-world CSV files. We then generate responses to these instructions with the agent framework and GPT-4. By training open-source LLMs on this dataset, we further develop DAAgent, specialized agents focused on data analysis. The largest model DAAgent-34B, achieves a better performance over GPT-3.5 by 3.9\%. To summarize, our main contributions are:
\begin{itemize}
    \item We propose InfiAgent-DABench, which is the first benchmark, to our knowledge, for evaluating agents on data analysis tasks. 
    \item We benchmark 34 LLMs and found that even state-of-the-art LLMs still have challenges handling data analysis tasks well. 
    \item We propose an instruction-tuning dataset DAInstruct for data analysis tasks and develop DAAgent, an open-source data analysis agent, which achieves a better performance over GPT-3.5 by 3.9\% on data analysis tasks.
\end{itemize}

\section{InfiAgent-DABench Benchmark}
In this section, we present InfiAgent-DABench, a benchmark created to assess the performance of LLM-based agents on data analysis tasks. Specifically, we design two main components of the benchmark: DAEval, the evaluation dataset which contains closed-form questions on data analysis, and the agent framework, which allows LLMs to solve data analysis problems in an agent setting. To ensure the quality of DAEval, we conduct a comprehensive human assessment and filter all unqualified data samples. 

\subsection{Dataset Construction}
Similar to other recent grounding dataset, such as VQA \citep{goyal2017making} and DocQA \citep{mathew2021docvqa}, DAEval is composed of realistic CSV files and the corresponding questions. The questions are generated according to several key concepts in the field of data analysis, which were identified through expert interviews. All questions are closed-form with corresponding answers for a straightforward and model-free assessment. Figure \ref{fig:workflow dataset construction} illustrates the whole process for constructing the dataset. We introduce the main steps as follows:

\textbf{Files Collection.}
We gather real-world CSV files from GitHub to form our dataset. 
We further assess the suitability of the CSV files for question generation based on the following criteria: 1. Meaningfulness: The presence of meaningful scenarios, 2. Language uniqueness: Predominantly English used in headers and content, 3. Scale sufficiency: The column count should be greater than 5 with a minimum row count of 20. 

\textbf{Description Generation.}
As directly putting all the content of the file into the LLMs to generate questions is usually infeasible, we utilize GPT-3.5 to summarize the following information for each CSV file: 1. Description of column names and contents, 2. Data types, 3. Identification of missing values, as a mediate step before question generation. More detailed information is in Appendix \ref{appendix:description generation}.

\textbf{Concepts Determination.}
To ensure that our dataset aligns with real-world demands, we conduct expert interviews to identify key concepts in data analysis. All the experts are professional data analysts. We list these concepts and corresponding explanations in Appendix \ref{appendix:concepts and explanations}. We remove data visualization from original concepts since we assume that a question requiring visualization can hardly be closed-form. We have a more detailed discussion on it in Appendix \ref{sec:limitations}. These concepts are used to guide the generation of questions in subsequent stages.

\textbf{Open-ended Question Generation.}
In this step, we leverage GPT-4 to generate open-ended questions based on both file descriptions and the pre-defined concepts. Each question involves one or multiple concepts. We prompt GPT-4 to generate questions with an increasing difficulty and more concepts for each file description. All questions for one file are generated in one dialogue to make sure no repeated questions. Prompts can be found in Appendix \ref{appendix:question generation}.

\textbf{Constraints and Format Requirements Generation.}
Considering closed-form questions can be easily evaluated through regular expression and exact match without any external models or human efforts, we introduce a format-prompting technique that prompts GPT-4 to generate detailed constraints and format requirements for open-ended questions to make them closed-form. We have noticed that most powerful LLMs like GPT-4 have potential for automatic evaluation for open-ended questions \citep{fu2023gptscore, chan2023chateval}, while many works point out issues of this way \citep{wang2023chatgpt, wang2023large}. We also conduct a pilot experiment on GPT-4 auto-evaluating for above open-ended questions. The results show that GPT-4 could only achieve 67\% consistency with human experts, indicating that GPT-4 are still unsatisfactory as an evaluator on data analysis tasks. Details about pilot experiment is in Appendix \ref{appendix:pilot}.

However, close-form questions allow us to bypass these issues. Constraints mainly contain detailed strict on methods, workflow or statistical parameters used to answer the question. Format requirements are series of ``@answer\_name[answer]'' where ``answer\_name'' and ``answer'' are automatically defined according to the specific question, also natural description of the range of value to each ``answer''. We also call each ``@answer\_name[answer]'' a subquestion in the following. We conduct human assessment and filtering to ensure adding constraints and format requirements on open-ended questions can make answers unique and easy to parse for evaluation. See Section \ref{sec:human_assessment} for details. Full prompts can be found in Appendix \ref{appendix:Constraints and Format requirements Generation}.  

\textbf{Response Gathering.}
We leverage OpenAI ADA to generate answers of closed-form questions. Each question is sent to ADA for 3 times. Only the answer is kept when ADA consistently generates the same answer in all trials. 

Figure \ref{fig:workflow dataset construction} showcases a question generated from the file \texttt{test\_ave.csv}. We provide an extended case study on generated questions in Appendix \ref{appendix:case study on generated questions}.

\subsection{Agent Framework}
We build an agent framework for LLM-based agents to reason in a ReAct \citep{yao2023react} way, interact with files and invoke tools such as a Python code sandbox for easy evaluation. Agents here can be a model deployed online, such as GPT-4, or local models based on vLLM \citep{kwon2023efficient}. Simultaneously, we develop a Docker-based local Python code sandbox that allows the isolated execution of code, avoiding safety issues.

\subsection{Human Assessment}
\label{sec:human_assessment}
To ensure a high-quality dataset, we conduct an in-depth human assessment. We invite experts to assess the dataset quality, and also conduct a comparison between human-made and GPT-4 generated data analysis questions.

\textbf{Metrics.} We evaluate the quality of files, questions and labels. For each part, we define one or more metrics which are crucial for a data analysis dataset. Here's a brief introduction:
\begin{itemize}
    \item \textbf{Suitableness}: This measures whether a CSV file is suitable for data analysis.

    \item \textbf{Reasonableness}: This measures whether the question, constraints and format requirements express in a natural and reasonable way, with no conflict with each other.

    \item \textbf{Value}: This measures the value of generated questions. We want the questions to be valuable in practice. For instance, calculating the mean of salaries could be valuable as a data analysis question, while calculating the mean of ID numbers is usually valueless.

    \item \textbf{Restrictiveness}: This measures whether the constraints are strict enough to ensure the uniqueness of answers.

    \item \textbf{Alignment}: This measures the alignment between questions and data file. Questions should only require data existed in the file, also the type and range of the data are aligned with questions.

    \item \textbf{Correctness}: This measures whether the labels for questions are correct.
\end{itemize}

\subsubsection{Dataset Quality}
We invite human experts to assess the quality of generated dataset and remove samples unqualified in any metric. We refer readers to Appendix \ref{appendix:questionnaire} for detailed settings for human assessment on dataset quality. 

\textbf{Results.} Table \ref{tab:human_assessment} shows the percentage of samples passed in specific metrics and overall pass rate. 85\% samples are qualified and kept in final, demonstrating the effectiveness of our dataset construction method. In addition, we conduct bad case analysis on the dataset in Appendix \ref{appendix:dataset error analysis}.

\begin{table*}
\centering
\footnotesize
\begin{tabular}{c|cccc|c|c}
\toprule
\multicolumn{1}{c}{\textbf{File}} & \multicolumn{4}{c}{\textbf{Question}} & \multicolumn{1}{c}{\textbf{Label}} & \multicolumn{1}{c}{} \\
\textbf{Suitableness} $ \uparrow $ & \textbf{Reasonableness} $ \uparrow $ & \textbf{Value} $ \uparrow $ & \textbf{Restrictiveness} $ \uparrow $ & \textbf{Alignment} $ \uparrow $ & \textbf{Correctness} $ \uparrow $ & \textbf{Overall} $ \uparrow $ \\
\midrule
0.98 & 0.95 & 0.98 & 0.94 & 0.98 & 0.93 & 0.85 \\
\bottomrule
\end{tabular}
\caption{\textbf{Results of human assessment on the quality of DAEval.} We conduct quality assessment over files, questions and labels, where each part has one or more metrics. We report the percentage of samples considered qualified on each metric. We keep questions qualified in all metrics. Overall represents the percentage of samples qualified in total.}
\label{tab:human_assessment}
\end{table*}

\begin{table*}
\centering
\footnotesize
\begin{tabular}{lcccc|c}
\toprule
\textbf{Source} & \textbf{Reasonableness} $ \uparrow $ & \textbf{Value} $ \uparrow $ & \textbf{Restrictiveness} $ \uparrow $ & \textbf{Alignment} $ \uparrow $ & \textbf{Difficulty (\%) }$ \uparrow $ \\ 
\midrule
Human & 4.66 & 4.73 & 4.14 & 4.71 & 39.34\\
GPT-4 & 4.70 & 4.85 & 4.74 & 4.92 & 60.66\\
\bottomrule
\end{tabular}
\caption{\textbf{Results of human assessment to compare questions generated from human and GPT-4.}  Reasonableness, Value, Restrictiveness and Alignment are graded from 1-5 (1 is worst and 5 is best). Difficulty is calculated by win rate between human and GPT-4.}
\label{tab:human_vs_gpt}
\vspace{-10pt}
\end{table*}

\subsubsection{Human vs GPT-4}
To ensure GPT-4 generated questions have a comparable quality with human-made ones, we invite experts to write data analysis questions. Then other experts are asked to grade from 1 to 5 (1 is worst and 5 is best) to metrics on questions mentioned above each. Additionally, we ask experts to decide which question is more difficult in a question pair randomly sampled from human-made and GPT-4 generated questions respectively to compare the difficulty of questions made from two sides. We calculate win rate to reflect the relative difficulty of questions. Detailed settings for this part of human assessment are in Appendix \ref{tab:human_vs_gpt}.

\textbf{Results.} Table \ref{tab:human_vs_gpt} shows results compared between human-made and GPT-4 generated questions. We found that GPT-4 generated questions take advantage on Restrictiveness, Alignment and Difficulty, and a slightly better performance on Reasonableness and Value, which indicates that DAEval are comparable with human-made dataset.

\subsection{Benchmark Statistics}
In this section, we provide statistics on DAEval. We split the dataset into validation set and test set. The validation set is open to public, including 257 questions with 52 csv files and the rest is for test set which is closed for avoiding data leakage. All the subsequent information is based on the validation set.

\textbf{Basic Statistics.} Table \ref{tab:basic_statistics} lists basic statistical information on files and questions, including the number of files, questions, subquestions, etc. 
\begin{table}[ht]
\centering
\begin{tabular}{l|c}
\toprule
\textbf{Dataset Component} & \textbf{Count} \\
\midrule
\# of CSV Files & 52 \\
\midrule
\# of Avg. Rows per File & 9639.7 \\
\midrule
\# of Avg. Columns per File & 10.6 \\
\midrule
\# of Questions & 257 \\
\midrule
\# of Subquestions & 461 \\
\midrule
\# of Max Subquestions per Question & 8 \\
\bottomrule
\end{tabular}
\caption{Basic statistics on the validation set of DAEval.}
\label{tab:basic_statistics}
\end{table}

\begin{figure}[htbp]
  \centering
  \includegraphics[width=0.8\linewidth]{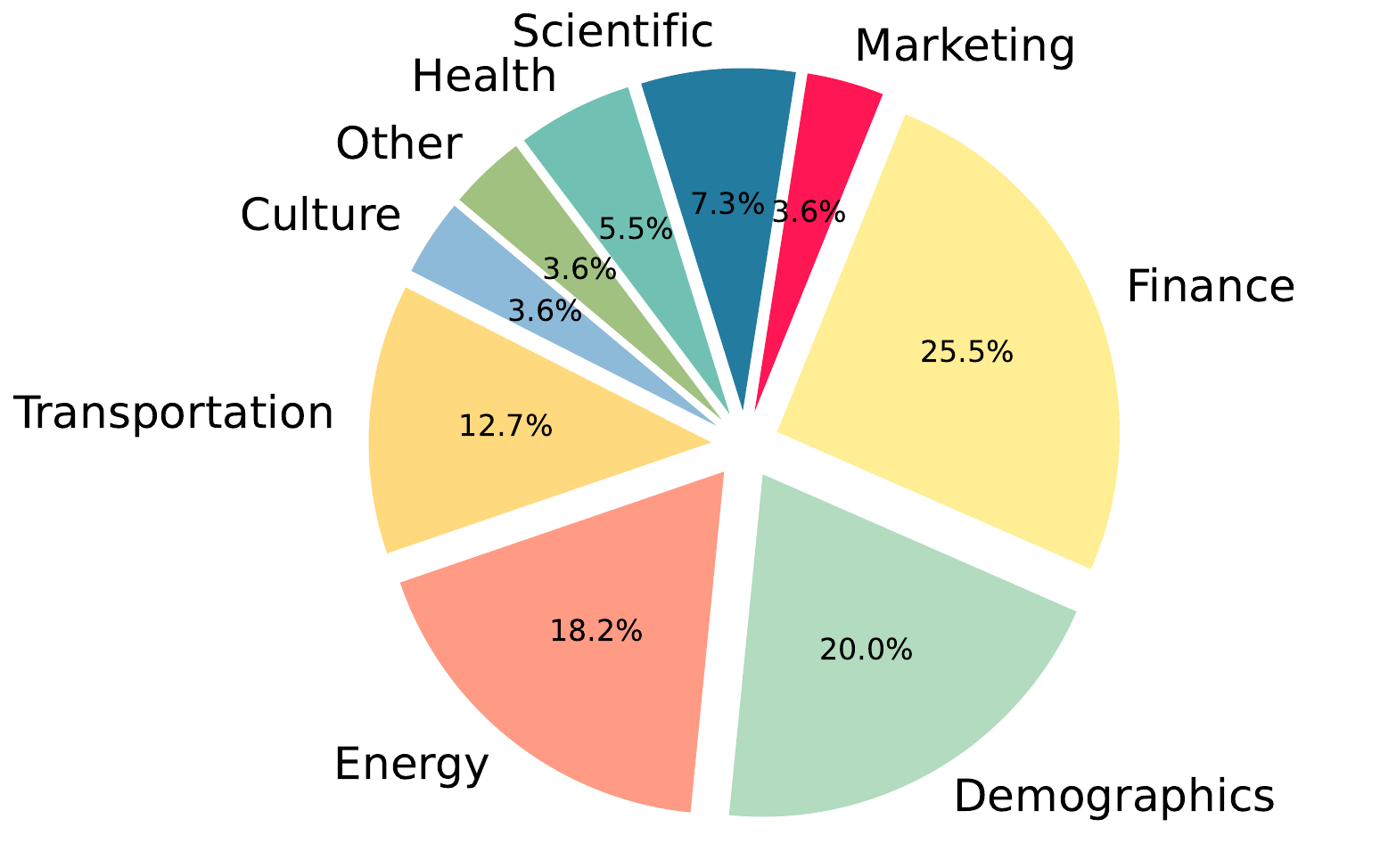}
  \caption{\textbf{Distribution of domains the files in.} For brevity, we use the first word of each category as the label. The completed domain names are listed in Appendix \ref{appendix:domains}.}
  \label{fig:domain}
\end{figure}

\textbf{Domains of Files.} 
Figure \ref{fig:domain} shows the distribution of different domains the files in. We cluster the files into 9 domains with the help of GPT-3.5. The result shows that our dataset covers a large range of domains. Details about the domains and corresponding descriptions are listed in Appendix \ref{appendix:domains}.

\begin{figure}[htbp]
  \centering
  \includegraphics[width=0.75\linewidth]{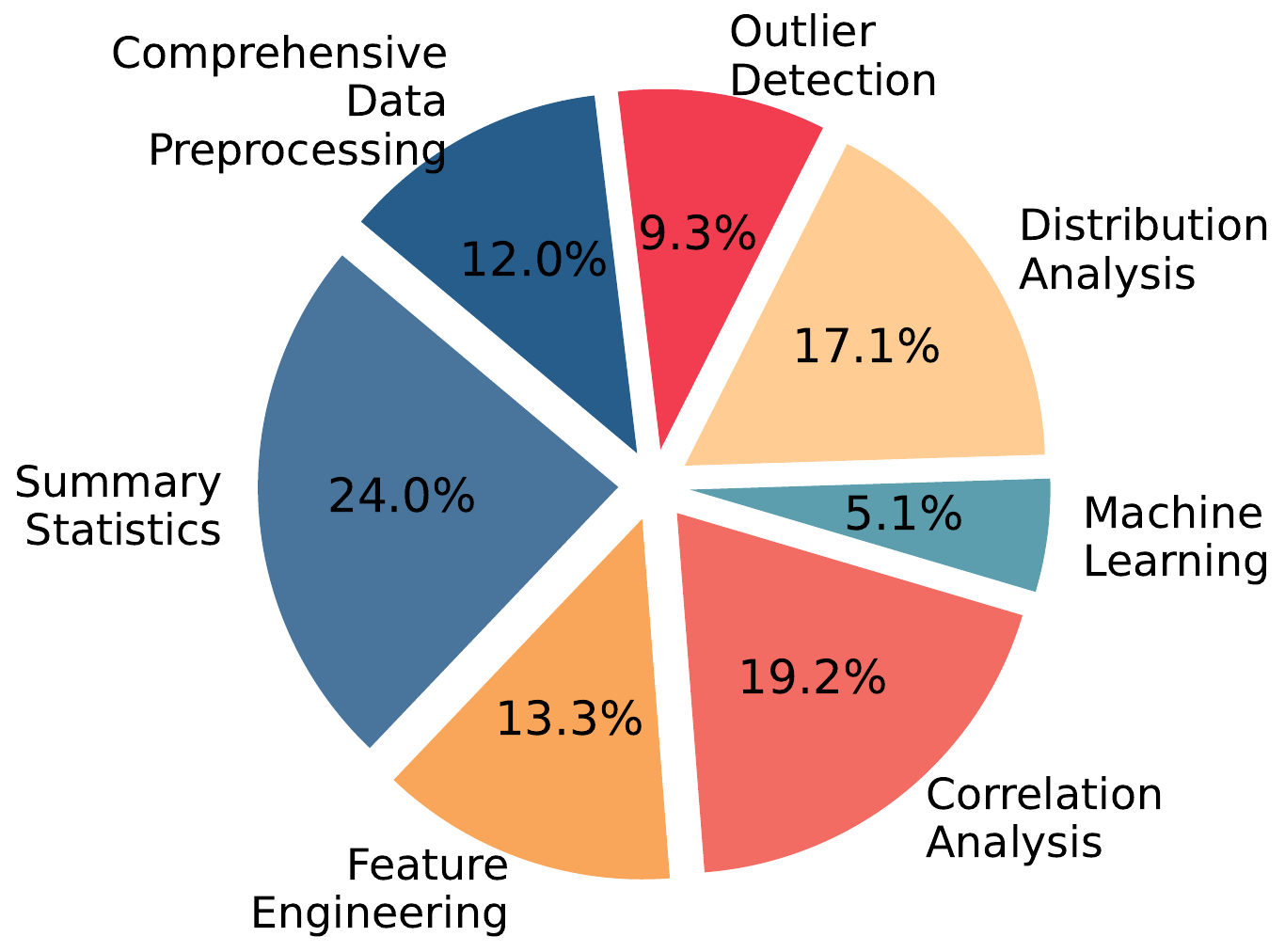}
  \caption{Distribution of concepts the questions involve.}
  \label{fig:concepts}
\end{figure}

\textbf{Concepts of Questions.}
Figure \ref{fig:concepts} shows the distribution of different concepts the questions involve. 
Table \ref{tab:question_difficulty_distribution} shows the distribution of questions based on the number of concepts involved. It reveals that a larger proportion of questions involve 1 or 2 concepts. This is reasonable that it's uncommon for a scenario where a question can involve so many concepts at once.

\begin{table*}[h]
\centering
\small
\begin{tabular}{llcccc}
\toprule
 &  & Easy & Medium & Hard & Total \\
\midrule
Concepts & Summary Statistics & 41 & 17 & 32 & 90 \\
 & Feature Engineering & 3 & 14 & 33 & 50 \\
 & Correlation Analysis & 10 & 32 & 30 & 72 \\
 & Machine Learning & 0 & 0 & 19 & 19 \\
 & Distribution Analysis & 21 & 23 & 20 & 64 \\
 & Outlier Detection & 5 & 20 & 10 & 35 \\
 & Comprehensive Data Preprocessing & 8 & 9 & 28 & 45 \\
 \midrule
 \# of Concepts per Question & 1 concept & 77 & 59 & 14 & 150 \\
 & 2 concepts & 4 & 28 & 65 & 97 \\
 & 3 concepts & 1 & 0 & 8 & 9 \\
 & 4 concepts & 0 & 0 & 1 & 1 \\
\midrule
Total &  & 82 & 87 & 88 & 257 \\
\bottomrule
\end{tabular}
\caption{Distribution of question difficulties across different concepts and number of concepts.}
\label{tab:question_difficulty_distribution}
\end{table*}

\textbf{Difficulty of Questions.} GPT-4 is utilized for the quantitative assessment of question difficulty in DAEval. Questions are classified into three levels: easy, medium, and hard. Table \ref{tab:question_difficulty_distribution} presents the distribution of these levels across different concepts and the number of concepts per question. The detailed method of quantitative assessment is described in Appendix \ref{appendix:methodology}. This indicates a wide spectrum of complexities within DAEval, ensuring a comprehensive evaluation.

\subsection{Instruction-tuning Dataset}
In this section, we introduce DAInstruct, an instruction-tuning dataset for data analysis, on which we train DAAgent, a specialized agent for data analysis. The collection of DAInstruct is carried out in three steps: 1. Creating a diverse set of data analysis questions for CSV files 2. Collecting response trajectories to these questions through our agent framework 3. Filtering low-quality samples with rules and GPT-4. Statistics of DAInstruct are in Appendix \ref{appendix:statistics on DAInstruct}. Moreover, a case study on DAInstruct is provided in Appendix \ref{appendix:case study on DAInstruct}. The following sections will detail these steps.

\textbf{Question Generation.} We employ GPT-4 to generate a set of keywords relevant to data analysis. Then, these keywords undergo a manual screening process to ensure relevance and accuracy. Thenm, GPT-4 is prompted with file description similar to those in DAEval, and above keywords to generate questions for each CSV file. The keywords and prompt can be found in Appendix \ref{appendix:prompts_for_DAInstruct_construction}. 

\textbf{Response Collection.} In this step, we leverage our agent framework to generate responses in a ReAct \citep{yao2023react} way. It is an iteration of problem-solving steps, where GPT-4 engages with code execution environment in a conversational format. The process includes reasoning, code writing, execution, and iterative refinement based on the results obtained. Details can be found in Appendix \ref{appendix:react_process}.

\textbf{Filtering.} This step involves a filtering with rules and GPT-4. Specific rules are employed to identify and eliminate data with hallucination and incomplete solution. Then, the data undergoes filtering with the aid of GPT-4. The criteria includes five aspects: Python code relevance, interpretation of system notifications, conciseness, understanding of the Python sandbox environment, and solution provision. All prompts can be found in Appendix \ref{appendix:prompts_for_DAInstruct_construction}.

\section{Experiments}
\begin{figure*}[t]
  \centering
  \includegraphics[width=\textwidth]{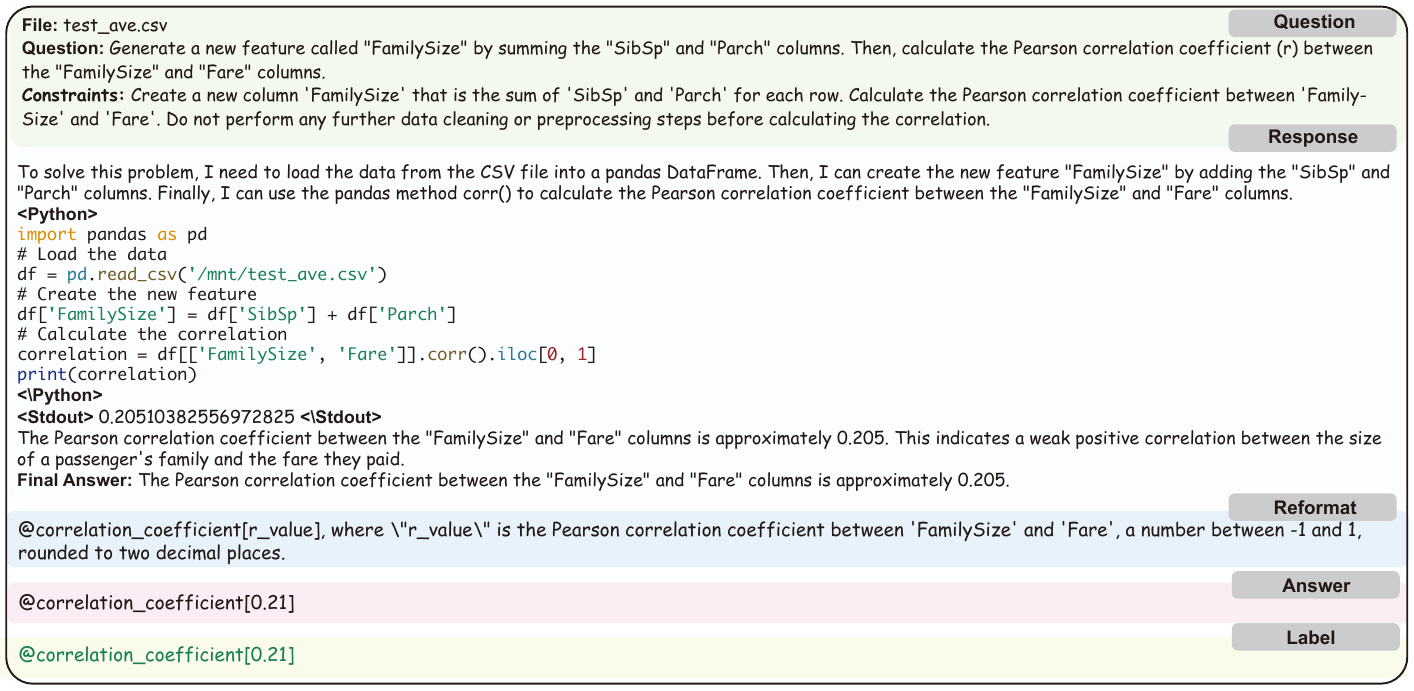}
  \vspace{-10pt}
  \caption{\textbf{The evaluation process.} We first prompt the agent with the file, the question and constraints. The agent answers with the calling of external Python sandbox. Then, we use GPT-3.5 to reformat the response into format requirements. At last, we use regular expression matching to draw the answer and compare it with the label.}
  \vspace{-10pt}
  \label{fig:evaluation}
\end{figure*}

\subsection{Setup}
\subsubsection{Models}
We categorize the models benchmarked by 4 groups: proprietary models, open-source general LLMs, open-source code LLMs and agent frameworks. 

\textbf{Proprietary Models.} This group consists of LLMs available as commercial services, providing API access for integration into various applications. We benchmark OpenAI GPT-4 \citep{openai2023gpt4}, GPT-3.5 \citep{openai2023gpt3.5}, Claude-2.1 \cite{claude2023}, Abab5.5 \citep{minimax2023abab5.5} and Gemini-Pro \citep{team2023gemini}.

\textbf{Open-source General LLMs.} This category includes open-source LLMs designed for a wide range of tasks and versatility in applications. We benchmark 
Vicuna \citep{vicuna2023}, ChatGLM \citep{zeng2022glm}, Baichuan \citep{yang2023baichuan}, Qwen \citep{bai2023qwen}, InternLM \citep{2023internlm}, AgentLM \citep{zeng2023agenttuning}, Mistral \citep{mistral2023} and Yi \citep{2023yi}.

\textbf{Open-source Code LLMs.} These are specialized LLMs tailored for understanding and generating programming code. Our experiments cover Code Llama \citep{rozière2023code}, WizardCoder \citep{luo2023wizardcoder}, Phind CodeLlama \citep{phind2023phindcodellama}, DeepSeek Coder \citep{deepseek-coder} and XwinCoder \citep{xwin-lm}.

\textbf{Agent Frameworks.} These frameworks contain one or more LLM-based agents which can reason, plan and use tools. We conduct evaluations on single agent framework XAgent \citep{xagent2023}, Qwen-Agent \citep{bai2023qwen} and multi-agents framework AutoGen \citep{wu2023autogen}. We use GPT-4 as the LLMs for XAgent and AutoGen, and Qwen-72B-Chat for Qwen-Agent.

\begin{figure*}[t]
  \centering
  \includegraphics[width=\textwidth]{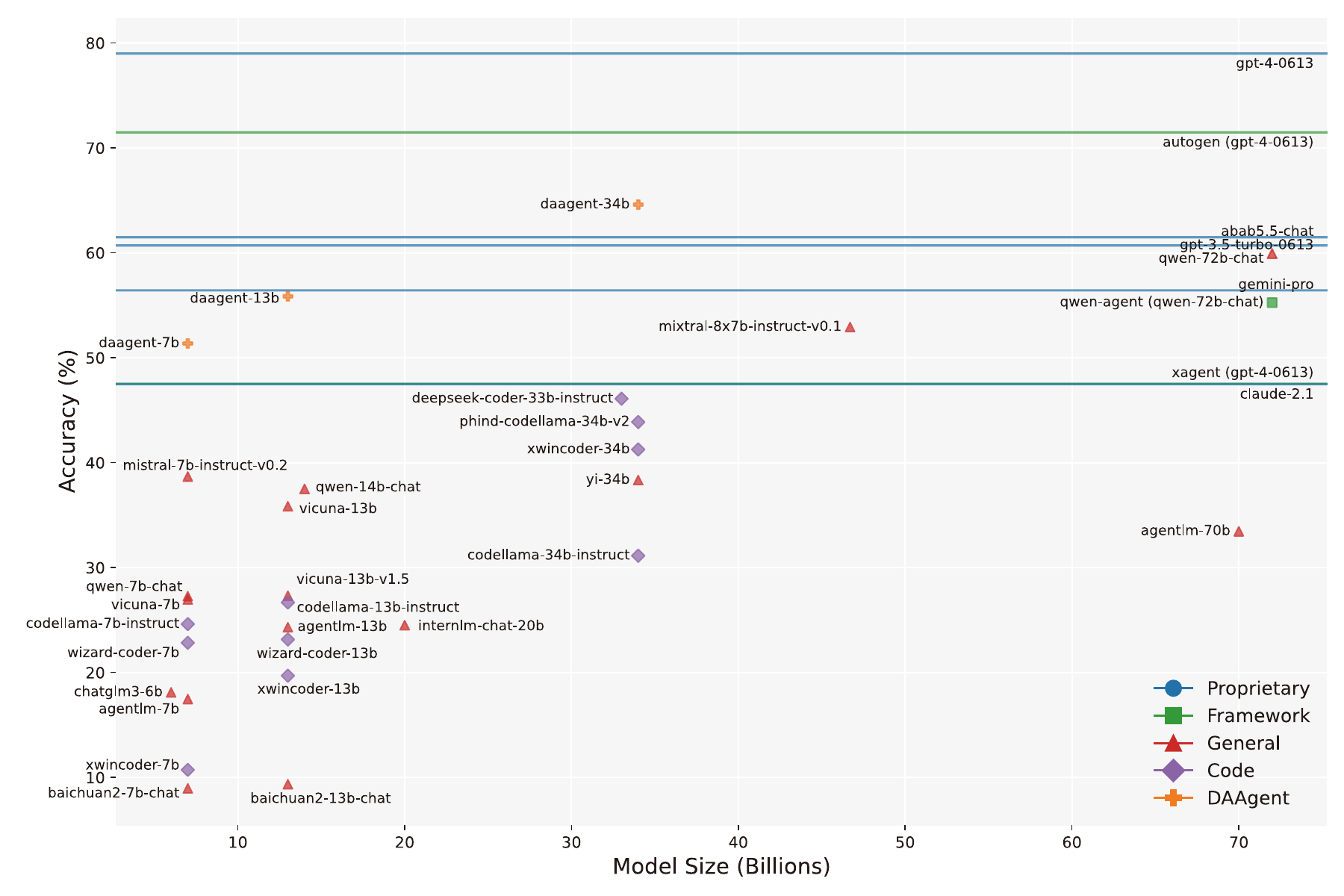}
  \caption{\textbf{The performance of models benchmarked in InfiAgent-DABench.} Models are categorized into 4 groups: Proprietary Models, Open-source General LLMs, Open-source Code LLMs, Agent Frameworks. LLMs behind agent frameworks are shown in parentheses. We also report our DAAgents. Models of known size are plotted as dots while models of unknown size as horizontal lines.}
  \label{fig:main_results}
  \vspace{-10pt}
\end{figure*}

\subsubsection{Implementation Details}
Figure \ref{fig:evaluation} illustrates a case on the process of evaluation. All experiments are based on our agent frameworks except those ready-made frameworks. We first prompt LLMs with the file, question and constraints. Considering that most models hardly follow the format requirements exactly, we add a reformat step after the original responses which formats the responses with the format requirements with one-shot prompting. We conduct ablation studies and manual check on this reformat step, indicating this step doesn't effect to the precision of evaluation. More details about reformatting can be found in Appendix \ref{appendix:reformat}. Other implementation details are in Appendix \ref{appendix:implementations}.

\subsection{Results}
Figure \ref{fig:main_results} shows performances and model sizes of different models benchmarked on validation set of DAEval. We provide precise performances in Appendix \ref{appendix:performances_on_dabench}.

\subsection{Findings}
Based on above results, we have several key findings:

\textbf{Data analysis tasks are challenging for current LLMs.} Among all benchmarked models, GPT-4 outperforms others with an accuracy of 78.99\%, highlights a significant room for improvement in even the most advanced models, since the upper limit of performance in DAEval is 100\%. Notably, the highest-performing open-source LLMs Qwen-72B-Chat, still lags behind by 19\% with GPT-4. The substantial gap between proprietary and open-source models indicates an overarching need for enhancement in open-source LLMs for data analysis tasks.

\textbf{Most powerful open-source LLMs achieve comparable performance with GPT-3.5.} We found that, open-source LLMs enjoy a fast development over past periods. Although early open-source models seem an obvious gap behind proprietary models, they catch up quickly. Qwen-72B-Chat achieves 59.92\%, the best performance among open-source models, which is only slightly lower than GPT-3.5 by 0.78\% and even better than Gemini-Pro by 3.50\%. 

\textbf{DAAgent achieves a better performance than GPT-3.5.} DAAgent-34B, our biggest model instruction-tuned by DAInstruct from CodeLlama, achieves a better performance with GPT-3.5 by 3.89\%, although it has much less parameters than that proprietary model. Otherwise, Table \ref{tab:daagents_with_different_sizes} compares DAAgent with corresponding size of Code Llama with general instruction-tuning showcases a large performance increase brought by instruction-tuning on DAInstruct. Both indicate DAInstruct can be seen as an effective instruction-tuning dataset to increase model's ability on data analysis. 

\begin{table}
\centering
\footnotesize
\begin{tabular}{lccc}
\toprule
\multicolumn{1}{c}{} & \multicolumn{3}{c}{\textbf{Accuracy (\%)}} \\
\textbf{Model} & 7B & 13B & 34B \\
\midrule
CodeLlama-Instruct & 24.61 & 26.67 & 31.13 \\ 
DAAgent & 51.36 & 55.86 & 64.59 \\
\bottomrule
\end{tabular}
\caption{The performance of models with different instruction-tuning dataset acorss all Code Llama model sizes.}
\label{tab:daagents_with_different_sizes}
\vspace{-10pt}
\end{table}

\section{Related works}
The rapid advancement of LLM-based agents necessitates precise and all-encompassing evaluation methods, while remains a challenge at present. Recent studies have introduced several new settings to evaluate from different angles. WebShop \citep{yao2022webshop} and WebArena \citep{zhou2023webarena} benchmark LLM-based agents in web environments. Some other benchmarks evaluate LLM-based agents with tool using, like ToolBench \citep{qin2023toolllm} and Gentopia \citep{xu2023gentopia}. In our work, we propose the InfiAgent-DABench that evaluates agents on data analysis tasks. Such tasks have a higher requirement for LLMs to produce structured and precise intermediate executable steps. 
Another series of works related to InfiAgent-DABench are Code LLMs and corresponding evaluations. Code LLMs, which specialize in understanding and generating programming code, have developed fast in the past few years~\citep{zan-etal-2023-large, zhang2023unifying, chen2021evaluating, rozière2023code, luo2023wizardcoder}. Various benchmarks have also been proposed, \emph{e.g.}, CodeXGLUE \citep{lu2021codexglue} and DS-1000 \citep{lai2023ds}. However, these benchmarks still miss a comprehensive evaluation to check agents end-to-end task solving capability. More details about Code LLMs and related benchmarks can be found in Appendix \ref{appendix:more_related_works}.

\section{Conclusion}
We introduce InfiAgent-DABench, the first benchmark specifically designed for evaluating LLM-based agents on data analysis tasks. Our extensive benchmarking of 34 cutting-edge LLMs reveals that contemporary models still face challenges in effectively managing data analysis tasks. Additionally, we develop DAAgent, a specialized agent for data analysis, trained on DAInstruct, an instrcution-tuning dataset generated by GPT-4. Our contributions not only offer a valuable benchmark for assessing LLM-based agents in data analysis but also provide insights into the current capabilities and limitations of these models in this domain.

\section*{Acknowledgements}
We appreciate the contribution of Haotian Zhou on helping construct the agent pipeline and environment. We also extend our gratitude to Rufei Liu, Feiyu Zhang and Xinxin Wang's hard work on data annotation.
\newpage

\bibliography{custom}

\begin{thebibliography}{68}
\providecommand{\natexlab}[1]{#1}
\providecommand{\url}[1]{\texttt{#1}}
\expandafter\ifx\csname urlstyle\endcsname\relax
  \providecommand{\doi}[1]{doi: #1}\else
  \providecommand{\doi}{doi: \begingroup \urlstyle{rm}\Url}\fi

\bibitem[01ai(2023)]{2023yi}
01ai.
\newblock Yi.
\newblock \url{https://01.ai/}, 2023.

\bibitem[Anthropic(2023)]{claude2023}
Anthropic.
\newblock Claude-2.1.
\newblock \url{https://www.anthropic.com/index/claude-2-1}, 2023.

\bibitem[Austin et~al.(2021)Austin, Odena, Nye, Bosma, Michalewski, Dohan, Jiang, Cai, Terry, Le, et~al.]{austin2021program}
Austin, J., Odena, A., Nye, M., Bosma, M., Michalewski, H., Dohan, D., Jiang, E., Cai, C., Terry, M., Le, Q., et~al.
\newblock Program synthesis with large language models.
\newblock \emph{arXiv preprint arXiv:2108.07732}, 2021.

\bibitem[Bai et~al.(2023)Bai, Bai, Chu, Cui, Dang, Deng, Fan, Ge, Han, Huang, et~al.]{bai2023qwen}
Bai, J., Bai, S., Chu, Y., Cui, Z., Dang, K., Deng, X., Fan, Y., Ge, W., Han, Y., Huang, F., et~al.
\newblock Qwen technical report.
\newblock \emph{arXiv preprint arXiv:2309.16609}, 2023.

\bibitem[Chan et~al.(2023)Chan, Chen, Su, Yu, Xue, Zhang, Fu, and Liu]{chan2023chateval}
Chan, C.-M., Chen, W., Su, Y., Yu, J., Xue, W., Zhang, S., Fu, J., and Liu, Z.
\newblock Chateval: Towards better llm-based evaluators through multi-agent debate.
\newblock \emph{arXiv preprint arXiv:2308.07201}, 2023.

\bibitem[Chen et~al.(2021)Chen, Tworek, Jun, Yuan, Pinto, Kaplan, Edwards, Burda, Joseph, Brockman, et~al.]{chen2021evaluating}
Chen, M., Tworek, J., Jun, H., Yuan, Q., Pinto, H. P. d.~O., Kaplan, J., Edwards, H., Burda, Y., Joseph, N., Brockman, G., et~al.
\newblock Evaluating large language models trained on code.
\newblock \emph{arXiv preprint arXiv:2107.03374}, 2021.

\bibitem[Chen et~al.(2023)Chen, Lin, Sch{\"a}rli, and Zhou]{chen2023teaching}
Chen, X., Lin, M., Sch{\"a}rli, N., and Zhou, D.
\newblock Teaching large language models to self-debug.
\newblock \emph{arXiv preprint arXiv:2304.05128}, 2023.

\bibitem[Chiang et~al.(2023)Chiang, Li, Lin, Sheng, Wu, Zhang, Zheng, Zhuang, Zhuang, Gonzalez, Stoica, and Xing]{vicuna2023}
Chiang, W.-L., Li, Z., Lin, Z., Sheng, Y., Wu, Z., Zhang, H., Zheng, L., Zhuang, S., Zhuang, Y., Gonzalez, J.~E., Stoica, I., and Xing, E.~P.
\newblock Vicuna: An open-source chatbot impressing gpt-4 with 90\%* chatgpt quality, March 2023.
\newblock URL \url{https://lmsys.org/blog/2023-03-30-vicuna/}.

\bibitem[Covington(2016)]{covington2016analytics}
Covington, D.
\newblock \emph{Analytics: Data Science, Data Analysis, and Predictive Analytics for Business}.
\newblock CreateSpace Independent Publishing Platform, 2016.

\bibitem[DeepSeek(2023)]{deepseek-coder}
DeepSeek.
\newblock Deepseek coder: Let the code write itself.
\newblock \url{https://github.com/deepseek-ai/DeepSeek-Coder}, 2023.

\bibitem[Fu et~al.(2023)Fu, Ng, Jiang, and Liu]{fu2023gptscore}
Fu, J., Ng, S.-K., Jiang, Z., and Liu, P.
\newblock Gptscore: Evaluate as you desire.
\newblock \emph{arXiv preprint arXiv:2302.04166}, 2023.

\bibitem[Goyal et~al.(2017)Goyal, Khot, Summers-Stay, Batra, and Parikh]{goyal2017making}
Goyal, Y., Khot, T., Summers-Stay, D., Batra, D., and Parikh, D.
\newblock Making the v in vqa matter: Elevating the role of image understanding in visual question answering, 2017.

\bibitem[Hardy \& Bryman(2004)Hardy and Bryman]{hardy2004handbook}
Hardy, M.~A. and Bryman, A.
\newblock Handbook of data analysis.
\newblock 2004.

\bibitem[Hendrycks et~al.(2021)Hendrycks, Basart, Kadavath, Mazeika, Arora, Guo, Burns, Puranik, He, Song, and Steinhardt]{hendrycks2021measuring}
Hendrycks, D., Basart, S., Kadavath, S., Mazeika, M., Arora, A., Guo, E., Burns, C., Puranik, S., He, H., Song, D., and Steinhardt, J.
\newblock Measuring coding challenge competence with apps, 2021.

\bibitem[Huang \& Chang(2023)Huang and Chang]{huang2023reasoning}
Huang, J. and Chang, K. C.-C.
\newblock Towards reasoning in large language models: A survey, 2023.

\bibitem[Kojima et~al.(2022)Kojima, Gu, Reid, Matsuo, and Iwasawa]{kojima2022large}
Kojima, T., Gu, S.~S., Reid, M., Matsuo, Y., and Iwasawa, Y.
\newblock Large language models are zero-shot reasoners.
\newblock \emph{Advances in neural information processing systems}, 35:\penalty0 22199--22213, 2022.

\bibitem[Koop(2022)]{koop2022analysis}
Koop, G.
\newblock \emph{Analysis of financial data}.
\newblock John Wiley \& Sons Inc., 2022.

\bibitem[Kwon et~al.(2023)Kwon, Li, Zhuang, Sheng, Zheng, Yu, Gonzalez, Zhang, and Stoica]{kwon2023efficient}
Kwon, W., Li, Z., Zhuang, S., Sheng, Y., Zheng, L., Yu, C.~H., Gonzalez, J.~E., Zhang, H., and Stoica, I.
\newblock Efficient memory management for large language model serving with pagedattention.
\newblock In \emph{Proceedings of the ACM SIGOPS 29th Symposium on Operating Systems Principles}, 2023.

\bibitem[Lai et~al.(2023)Lai, Li, Wang, Zhang, Zhong, Zettlemoyer, Yih, Fried, Wang, and Yu]{lai2023ds}
Lai, Y., Li, C., Wang, Y., Zhang, T., Zhong, R., Zettlemoyer, L., Yih, W.-t., Fried, D., Wang, S., and Yu, T.
\newblock Ds-1000: A natural and reliable benchmark for data science code generation.
\newblock In \emph{International Conference on Machine Learning}, pp.\  18319--18345. PMLR, 2023.

\bibitem[Li et~al.(2023)Li, Song, Yu, Yu, Li, Huang, and Li]{li2023api}
Li, M., Song, F., Yu, B., Yu, H., Li, Z., Huang, F., and Li, Y.
\newblock Api-bank: A benchmark for tool-augmented llms.
\newblock \emph{arXiv preprint arXiv:2304.08244}, 2023.

\bibitem[Li et~al.(2022)Li, Choi, Chung, Kushman, Schrittwieser, Leblond, Eccles, Keeling, Gimeno, Dal~Lago, et~al.]{li2022competition}
Li, Y., Choi, D., Chung, J., Kushman, N., Schrittwieser, J., Leblond, R., Eccles, T., Keeling, J., Gimeno, F., Dal~Lago, A., et~al.
\newblock Competition-level code generation with alphacode.
\newblock \emph{Science}, 378\penalty0 (6624):\penalty0 1092--1097, 2022.

\bibitem[Liang et~al.(2023)Liang, Wang, Huang, Wu, Wu, Lu, Ma, and Li]{liang2023unleashing}
Liang, X., Wang, B., Huang, H., Wu, S., Wu, P., Lu, L., Ma, Z., and Li, Z.
\newblock Unleashing infinite-length input capacity for large-scale language models with self-controlled memory system.
\newblock \emph{arXiv preprint arXiv:2304.13343}, 2023.

\bibitem[Liu et~al.(2023)Liu, Sarkar, Negreanu, Zorn, Williams, Toronto, and Gordon]{liu2023wants}
Liu, M.~X., Sarkar, A., Negreanu, C., Zorn, B., Williams, J., Toronto, N., and Gordon, A.~D.
\newblock “what it wants me to say”: Bridging the abstraction gap between end-user programmers and code-generating large language models.
\newblock In \emph{Proceedings of the 2023 CHI Conference on Human Factors in Computing Systems}, pp.\  1--31, 2023.

\bibitem[Lu et~al.(2021)Lu, Guo, Ren, Huang, Svyatkovskiy, Blanco, Clement, Drain, Jiang, Tang, et~al.]{lu2021codexglue}
Lu, S., Guo, D., Ren, S., Huang, J., Svyatkovskiy, A., Blanco, A., Clement, C., Drain, D., Jiang, D., Tang, D., et~al.
\newblock Codexglue: A machine learning benchmark dataset for code understanding and generation.
\newblock \emph{arXiv preprint arXiv:2102.04664}, 2021.

\bibitem[Lucas(2023)]{2023openinterpreter}
Lucas, K.
\newblock Open interpreter.
\newblock \url{https://github.com/KillianLucas/open-interpreter}, 2023.

\bibitem[Luo et~al.(2023)Luo, Xu, Zhao, Sun, Geng, Hu, Tao, Ma, Lin, and Jiang]{luo2023wizardcoder}
Luo, Z., Xu, C., Zhao, P., Sun, Q., Geng, X., Hu, W., Tao, C., Ma, J., Lin, Q., and Jiang, D.
\newblock Wizardcoder: Empowering code large language models with evol-instruct.
\newblock \emph{arXiv preprint arXiv:2306.08568}, 2023.

\bibitem[Mathew et~al.(2021)Mathew, Karatzas, and Jawahar]{mathew2021docvqa}
Mathew, M., Karatzas, D., and Jawahar, C.~V.
\newblock Docvqa: A dataset for vqa on document images, 2021.

\bibitem[MiniMax(2023)]{minimax2023abab5.5}
MiniMax.
\newblock Abab5.5.
\newblock \url{https://api.minimax.chat/}, 2023.

\bibitem[Mistral.ai(2023)]{mistral2023}
Mistral.ai.
\newblock Mistral.
\newblock \url{https://mistral.ai/product/}, 2023.

\bibitem[Nakajima(2023)]{2023babyagi}
Nakajima, Y.
\newblock Babyagi.
\newblock \url{https://github.com/yoheinakajima/babyagi}, 2023.

\bibitem[OpenAI(2023{\natexlab{a}})]{openai2023gpt3.5}
OpenAI.
\newblock Openai models - openai api.
\newblock \url{https://platform.openai.com/docs/models/gpt-3-5}, 2023{\natexlab{a}}.

\bibitem[OpenAI(2023{\natexlab{b}})]{openai2023gpt4}
OpenAI.
\newblock Gpt-4 technical report.
\newblock 2023{\natexlab{b}}.
\newblock URL \url{https://arxiv.org/pdf/2303.08774.pdf}.

\bibitem[Patil et~al.(2023)Patil, Zhang, Wang, and Gonzalez]{patil2023gorilla}
Patil, S.~G., Zhang, T., Wang, X., and Gonzalez, J.~E.
\newblock Gorilla: Large language model connected with massive apis.
\newblock \emph{arXiv preprint arXiv:2305.15334}, 2023.

\bibitem[Phind(2023)]{phind2023phindcodellama}
Phind.
\newblock Phind code llama.
\newblock \url{https://www.phind.com/blog/code-llama-beats-gpt4}, 2023.

\bibitem[Qiao et~al.(2023)Qiao, Li, Zhang, He, Kang, Zhang, Yang, Dong, Zhang, Wang, et~al.]{qiao2023taskweaver}
Qiao, B., Li, L., Zhang, X., He, S., Kang, Y., Zhang, C., Yang, F., Dong, H., Zhang, J., Wang, L., et~al.
\newblock Taskweaver: A code-first agent framework.
\newblock \emph{arXiv preprint arXiv:2311.17541}, 2023.

\bibitem[Qin et~al.(2023{\natexlab{a}})Qin, Hu, Lin, Chen, Ding, Cui, Zeng, Huang, Xiao, Han, Fung, Su, Wang, Qian, Tian, Zhu, Liang, Shen, Xu, Zhang, Ye, Li, Tang, Yi, Zhu, Dai, Yan, Cong, Lu, Zhao, Huang, Yan, Han, Sun, Li, Phang, Yang, Wu, Ji, Liu, and Sun]{qin2023tool}
Qin, Y., Hu, S., Lin, Y., Chen, W., Ding, N., Cui, G., Zeng, Z., Huang, Y., Xiao, C., Han, C., Fung, Y.~R., Su, Y., Wang, H., Qian, C., Tian, R., Zhu, K., Liang, S., Shen, X., Xu, B., Zhang, Z., Ye, Y., Li, B., Tang, Z., Yi, J., Zhu, Y., Dai, Z., Yan, L., Cong, X., Lu, Y., Zhao, W., Huang, Y., Yan, J., Han, X., Sun, X., Li, D., Phang, J., Yang, C., Wu, T., Ji, H., Liu, Z., and Sun, M.
\newblock Tool learning with foundation models, 2023{\natexlab{a}}.

\bibitem[Qin et~al.(2023{\natexlab{b}})Qin, Liang, Ye, Zhu, Yan, Lu, Lin, Cong, Tang, Qian, Zhao, Hong, Tian, Xie, Zhou, Gerstein, Li, Liu, and Sun]{qin2023toolllm}
Qin, Y., Liang, S., Ye, Y., Zhu, K., Yan, L., Lu, Y., Lin, Y., Cong, X., Tang, X., Qian, B., Zhao, S., Hong, L., Tian, R., Xie, R., Zhou, J., Gerstein, M., Li, D., Liu, Z., and Sun, M.
\newblock Toolllm: Facilitating large language models to master 16000+ real-world apis, 2023{\natexlab{b}}.

\bibitem[Reddy \& Aggarwal(2015)Reddy and Aggarwal]{reddy2015healthcare}
Reddy, C.~K. and Aggarwal, C.~C.
\newblock \emph{Healthcare data analytics}, volume~36.
\newblock CRC Press, 2015.

\bibitem[Reworkd(2023)]{2023agentgpt}
Reworkd.
\newblock Agentgpt.
\newblock \url{https://github.com/reworkd/AgentGPT}, 2023.

\bibitem[Rozière et~al.(2023)Rozière, Gehring, Gloeckle, Sootla, Gat, Tan, Adi, Liu, Remez, Rapin, Kozhevnikov, Evtimov, Bitton, Bhatt, Ferrer, Grattafiori, Xiong, Défossez, Copet, Azhar, Touvron, Martin, Usunier, Scialom, and Synnaeve]{rozière2023code}
Rozière, B., Gehring, J., Gloeckle, F., Sootla, S., Gat, I., Tan, X.~E., Adi, Y., Liu, J., Remez, T., Rapin, J., Kozhevnikov, A., Evtimov, I., Bitton, J., Bhatt, M., Ferrer, C.~C., Grattafiori, A., Xiong, W., Défossez, A., Copet, J., Azhar, F., Touvron, H., Martin, L., Usunier, N., Scialom, T., and Synnaeve, G.
\newblock Code llama: Open foundation models for code.
\newblock 2023.
\newblock URL \url{https://arxiv.org/pdf/2308.12950.pdf}.

\bibitem[Sabina \& Zalta(2020)Sabina and Zalta]{sabina2020scientific}
Sabina, L. and Zalta, E.~N.
\newblock Scientific research and big data.
\newblock \emph{The Stanford Encyclopedia of Philosophy (Summer 2020 Edition)}, 2020.

\bibitem[Sun et~al.(2023)Sun, Zhuang, Kong, Dai, and Zhang]{sun2023adaplanner}
Sun, H., Zhuang, Y., Kong, L., Dai, B., and Zhang, C.
\newblock Adaplanner: Adaptive planning from feedback with language models.
\newblock \emph{arXiv preprint arXiv:2305.16653}, 2023.

\bibitem[Team et~al.(2023)Team, Anil, Borgeaud, Wu, Alayrac, Yu, Soricut, Schalkwyk, Dai, Hauth, et~al.]{team2023gemini}
Team, G., Anil, R., Borgeaud, S., Wu, Y., Alayrac, J.-B., Yu, J., Soricut, R., Schalkwyk, J., Dai, A.~M., Hauth, A., et~al.
\newblock Gemini: a family of highly capable multimodal models.
\newblock \emph{arXiv preprint arXiv:2312.11805}, 2023.

\bibitem[Team(2023{\natexlab{a}})]{2023internlm}
Team, I.
\newblock Internlm: A multilingual language model with progressively enhanced capabilities.
\newblock \url{https://github.com/InternLM/InternLM}, 2023{\natexlab{a}}.

\bibitem[Team(2023{\natexlab{b}})]{xagent2023}
Team, X.
\newblock Xagent: An autonomous agent for complex task solving, 2023{\natexlab{b}}.

\bibitem[Team(2023{\natexlab{c}})]{xwin-lm}
Team, X.-L.
\newblock Xwin-lm, 9 2023{\natexlab{c}}.
\newblock URL \url{https://github.com/Xwin-LM/Xwin-LM}.

\bibitem[Torantulino(2023)]{2023autogpt}
Torantulino.
\newblock Autogpt.
\newblock \url{https://github.com/Significant-Gravitas/AutoGPT}, 2023.

\bibitem[Wang et~al.(2023{\natexlab{a}})Wang, Liang, Meng, Shi, Li, Xu, Qu, and Zhou]{wang2023chatgpt}
Wang, J., Liang, Y., Meng, F., Shi, H., Li, Z., Xu, J., Qu, J., and Zhou, J.
\newblock Is chatgpt a good nlg evaluator? a preliminary study.
\newblock \emph{arXiv preprint arXiv:2303.04048}, 2023{\natexlab{a}}.

\bibitem[Wang et~al.(2023{\natexlab{b}})Wang, Ma, Feng, Zhang, Yang, Zhang, Chen, Tang, Chen, Lin, et~al.]{wang2023survey}
Wang, L., Ma, C., Feng, X., Zhang, Z., Yang, H., Zhang, J., Chen, Z., Tang, J., Chen, X., Lin, Y., et~al.
\newblock A survey on large language model based autonomous agents.
\newblock \emph{arXiv preprint arXiv:2308.11432}, 2023{\natexlab{b}}.

\bibitem[Wang et~al.(2023{\natexlab{c}})Wang, Li, Chen, Cai, Zhu, Lin, Cao, Liu, Liu, and Sui]{wang2023large}
Wang, P., Li, L., Chen, L., Cai, Z., Zhu, D., Lin, B., Cao, Y., Liu, Q., Liu, T., and Sui, Z.
\newblock Large language models are not fair evaluators, 2023{\natexlab{c}}.

\bibitem[Wei et~al.(2022)Wei, Wang, Schuurmans, Bosma, Xia, Chi, Le, Zhou, et~al.]{wei2022chain}
Wei, J., Wang, X., Schuurmans, D., Bosma, M., Xia, F., Chi, E., Le, Q.~V., Zhou, D., et~al.
\newblock Chain-of-thought prompting elicits reasoning in large language models.
\newblock \emph{Advances in Neural Information Processing Systems}, 35:\penalty0 24824--24837, 2022.

\bibitem[Wu et~al.(2023)Wu, Bansal, Zhang, Wu, Li, Zhu, Jiang, Zhang, Zhang, Liu, Awadallah, White, Burger, and Wang]{wu2023autogen}
Wu, Q., Bansal, G., Zhang, J., Wu, Y., Li, B., Zhu, E., Jiang, L., Zhang, X., Zhang, S., Liu, J., Awadallah, A.~H., White, R.~W., Burger, D., and Wang, C.
\newblock Autogen: Enabling next-gen llm applications via multi-agent conversation framework.
\newblock 2023.

\bibitem[Xi et~al.(2023)Xi, Chen, Guo, He, Ding, Hong, Zhang, Wang, Jin, Zhou, et~al.]{xi2023rise}
Xi, Z., Chen, W., Guo, X., He, W., Ding, Y., Hong, B., Zhang, M., Wang, J., Jin, S., Zhou, E., et~al.
\newblock The rise and potential of large language model based agents: A survey.
\newblock \emph{arXiv preprint arXiv:2309.07864}, 2023.

\bibitem[Xie et~al.(2023)Xie, Zhou, Cheng, Shi, Weng, Liu, Hua, Zhao, Liu, Liu, et~al.]{xie2023openagents}
Xie, T., Zhou, F., Cheng, Z., Shi, P., Weng, L., Liu, Y., Hua, T.~J., Zhao, J., Liu, Q., Liu, C., et~al.
\newblock Openagents: An open platform for language agents in the wild.
\newblock \emph{arXiv preprint arXiv:2310.10634}, 2023.

\bibitem[Xu et~al.(2023{\natexlab{a}})Xu, Liu, Shen, Han, Li, Yue, Peng, Liu, Yao, and Xu]{xu2023gentopia}
Xu, B., Liu, X., Shen, H., Han, Z., Li, Y., Yue, M., Peng, Z., Liu, Y., Yao, Z., and Xu, D.
\newblock Gentopia: A collaborative platform for tool-augmented llms, 2023{\natexlab{a}}.

\bibitem[Xu et~al.(2023{\natexlab{b}})Xu, Peng, Lei, Mukherjee, Liu, and Xu]{xu2023rewoo}
Xu, B., Peng, Z., Lei, B., Mukherjee, S., Liu, Y., and Xu, D.
\newblock Rewoo: Decoupling reasoning from observations for efficient augmented language models.
\newblock \emph{arXiv preprint arXiv:2305.18323}, 2023{\natexlab{b}}.

\bibitem[Yang et~al.(2023)Yang, Xiao, Wang, Zhang, Bian, Yin, Lv, Pan, Wang, Yan, Yang, Deng, Wang, Liu, Ai, Dong, Zhao, Xu, Sun, Zhang, Liu, Ji, Xie, Dai, Fang, Su, Song, Liu, Ru, Ma, Wang, Liu, Lin, Nie, Guo, Sun, Zhang, Li, Li, Cheng, Chen, Zeng, Wang, Chen, Men, Yu, Pan, Shen, Wang, Li, Jiang, Gao, Zhang, Zhou, and Wu]{yang2023baichuan}
Yang, A., Xiao, B., Wang, B., Zhang, B., Bian, C., Yin, C., Lv, C., Pan, D., Wang, D., Yan, D., Yang, F., Deng, F., Wang, F., Liu, F., Ai, G., Dong, G., Zhao, H., Xu, H., Sun, H., Zhang, H., Liu, H., Ji, J., Xie, J., Dai, J., Fang, K., Su, L., Song, L., Liu, L., Ru, L., Ma, L., Wang, M., Liu, M., Lin, M., Nie, N., Guo, P., Sun, R., Zhang, T., Li, T., Li, T., Cheng, W., Chen, W., Zeng, X., Wang, X., Chen, X., Men, X., Yu, X., Pan, X., Shen, Y., Wang, Y., Li, Y., Jiang, Y., Gao, Y., Zhang, Y., Zhou, Z., and Wu, Z.
\newblock Baichuan 2: Open large-scale language models, 2023.

\bibitem[Yang et~al.(2024)Yang, Liu, Wu, Yang, Fung, Li, Huang, Cao, Wang, Wang, et~al.]{yang2024if}
Yang, K., Liu, J., Wu, J., Yang, C., Fung, Y.~R., Li, S., Huang, Z., Cao, X., Wang, X., Wang, Y., et~al.
\newblock If llm is the wizard, then code is the wand: A survey on how code empowers large language models to serve as intelligent agents.
\newblock \emph{arXiv preprint arXiv:2401.00812}, 2024.

\bibitem[Yao et~al.(2022{\natexlab{a}})Yao, Chen, Yang, and Narasimhan]{yao2022webshop}
Yao, S., Chen, H., Yang, J., and Narasimhan, K.
\newblock Webshop: Towards scalable real-world web interaction with grounded language agents.
\newblock \emph{Advances in Neural Information Processing Systems}, 35:\penalty0 20744--20757, 2022{\natexlab{a}}.

\bibitem[Yao et~al.(2022{\natexlab{b}})Yao, Zhao, Yu, Du, Shafran, Narasimhan, and Cao]{yao2022react}
Yao, S., Zhao, J., Yu, D., Du, N., Shafran, I., Narasimhan, K., and Cao, Y.
\newblock React: Synergizing reasoning and acting in language models.
\newblock \emph{arXiv preprint arXiv:2210.03629}, 2022{\natexlab{b}}.

\bibitem[Yao et~al.(2023)Yao, Zhao, Yu, Du, Shafran, Narasimhan, and Cao]{yao2023react}
Yao, S., Zhao, J., Yu, D., Du, N., Shafran, I., Narasimhan, K., and Cao, Y.
\newblock React: Synergizing reasoning and acting in language models.
\newblock 2023.
\newblock URL \url{https://arxiv.org/pdf/2210.03629.pdf}.

\bibitem[Zan et~al.(2023)Zan, Chen, Zhang, Lu, Wu, Guan, Yongji, and Lou]{zan-etal-2023-large}
Zan, D., Chen, B., Zhang, F., Lu, D., Wu, B., Guan, B., Yongji, W., and Lou, J.-G.
\newblock Large language models meet {NL}2{C}ode: A survey.
\newblock In Rogers, A., Boyd-Graber, J., and Okazaki, N. (eds.), \emph{Proceedings of the 61st Annual Meeting of the Association for Computational Linguistics (Volume 1: Long Papers)}, pp.\  7443--7464, Toronto, Canada, July 2023. Association for Computational Linguistics.
\newblock \doi{10.18653/v1/2023.acl-long.411}.
\newblock URL \url{https://aclanthology.org/2023.acl-long.411}.

\bibitem[Zeng et~al.(2022)Zeng, Liu, Du, Wang, Lai, Ding, Yang, Xu, Zheng, Xia, et~al.]{zeng2022glm}
Zeng, A., Liu, X., Du, Z., Wang, Z., Lai, H., Ding, M., Yang, Z., Xu, Y., Zheng, W., Xia, X., et~al.
\newblock Glm-130b: An open bilingual pre-trained model.
\newblock \emph{arXiv preprint arXiv:2210.02414}, 2022.

\bibitem[Zeng et~al.(2023)Zeng, Liu, Lu, Wang, Liu, Dong, and Tang]{zeng2023agenttuning}
Zeng, A., Liu, M., Lu, R., Wang, B., Liu, X., Dong, Y., and Tang, J.
\newblock Agenttuning: Enabling generalized agent abilities for llms, 2023.

\bibitem[Zhang et~al.(2023{\natexlab{a}})Zhang, Zhao, Kang, and Liu]{zhang2023memory}
Zhang, K., Zhao, F., Kang, Y., and Liu, X.
\newblock Memory-augmented llm personalization with short-and long-term memory coordination.
\newblock \emph{arXiv preprint arXiv:2309.11696}, 2023{\natexlab{a}}.

\bibitem[Zhang et~al.(2023{\natexlab{b}})Zhang, Chen, Liu, Liao, Gong, Yu, Li, and Wang]{zhang2023unifying}
Zhang, Z., Chen, C., Liu, B., Liao, C., Gong, Z., Yu, H., Li, J., and Wang, R.
\newblock Unifying the perspectives of nlp and software engineering: A survey on language models for code, 2023{\natexlab{b}}.

\bibitem[Zhao et~al.(2023)Zhao, Jin, and Cheng]{zhao2023depth}
Zhao, P., Jin, Z., and Cheng, N.
\newblock An in-depth survey of large language model-based artificial intelligence agents.
\newblock \emph{arXiv preprint arXiv:2309.14365}, 2023.

\bibitem[Zhou et~al.(2023)Zhou, Xu, Zhu, Zhou, Lo, Sridhar, Cheng, Ou, Bisk, Fried, Alon, and Neubig]{zhou2023webarena}
Zhou, S., Xu, F.~F., Zhu, H., Zhou, X., Lo, R., Sridhar, A., Cheng, X., Ou, T., Bisk, Y., Fried, D., Alon, U., and Neubig, G.
\newblock Webarena: A realistic web environment for building autonomous agents, 2023.

\end{thebibliography}
\bibliographystyle{arxiv0301}

\appendix
\section{Limitations and Future work}
\label{sec:limitations}
A significant limitation of our benchmark is its exclusion of questions related to visualization. During expert interviews, we recognized the critical role of visualization in data analysis. However, designing closed-form questions for visualization is greatly challenging, as the outcomes are often pictorial rather than textual. To evaluate, two potential approaches emerge: 

\textbf{Evaluating code correctness.} This method allows for the use of text-only LLMs in evaluation. While it benefits from leveraging existing LLMs, this approach might lead to inaccuracies. The primary goal in data analysis tasks is to generate accurate visual representations, not just the underlying code. Only assessing from code could be challenging for a precise evaluation in some situations.

\textbf{Evaluating with multimodal models.} \citet{bai2023qwen} leverages vision-language models like GPT-4 (Vision) \citep{openai2023gpt4} as an evaluator. This approach directly assesses the accuracy of the visual outputs. While this method is advantageous in evaluating the end product, current multimodal models are suffering from suboptimal performance and costly, leading to unsatisfactory assessment of data visualization. 

Neither of above approaches can achieve the accuracy and convenience of closed-form assessments. We leave the precise evaluation of visualization in data analysis as a direction for future research.

\section{Detailed related works }
\label{appendix:more_related_works}
\subsection{Code LLMs}
Code LLMs are large language models specialized in understanding and generating programming code. These LLMs enjoy a fast development over past few years \citep{zan-etal-2023-large, zhang2023unifying}. Codex \citep{chen2021evaluating} is one of the most famous in these models provided by a commercial API from OpenAI. Meanwhile, researchers have developed a large sum of open-source code LLMs, \emph{e.g.}, Code Llama \citep{rozière2023code}, WizardCoder \citep{luo2023wizardcoder}, XwinCoder \citep{xwin-lm} and DeepSeek Coder \citep{deepseek-coder}. 

\subsection{Benchmarks for Code} 
Over the past years, various benchmarks with different tasks have been proposed to evaluate models for code. CodeXGLUE \citep{lu2021codexglue} is a comprehensive benchmark covering tasks for both code understanding and generation. In addition, some benchmarks focus on code generation, especially in a competition-level \citep{chen2021evaluating,austin2021program,hendrycks2021measuring,li2022competition}. DS-1000 \citep{lai2023ds} is a benchmark for code generation in data science. \citet{bai2023qwen} creates a dataset containing math, data visualization and other general-purpose tasks to assess LLMs' ability on using the Python code interpreter. We have a more detailed comparison between InfiAgent-DABench and this work in Appendix \ref{appendix:dabench_vs_qwen}.

\section{Comparison between InfiAgent-DABench and Code Interpreter Benchmark in Qwen-Agent}
\label{appendix:dabench_vs_qwen}
There are two main differences between InfiAgent-DABench and code interpreter benchmark in Qwen-Agent \citep{bai2023qwen}: 1. We benchmark LLMs in data analysis tasks covering series of concepts while \citet{bai2023qwen} only contains a few data analysis questions in data visualization and its other general-purpose tasks. 2. Our benchmark takes a closed-form evaluation on correctness which doesn't rely on any external models as evaluators. More importantly, \citet{bai2023qwen} evaluates code correctness for math and data visualization problems with the help of current multi-modal models but only code executability for its general-purpose tasks.

\section{Performances on InfiAgent-DABench}
\label{appendix:performances_on_dabench}
\begin{figure}[t]
  \centering
  \includegraphics[width=0.9\linewidth]{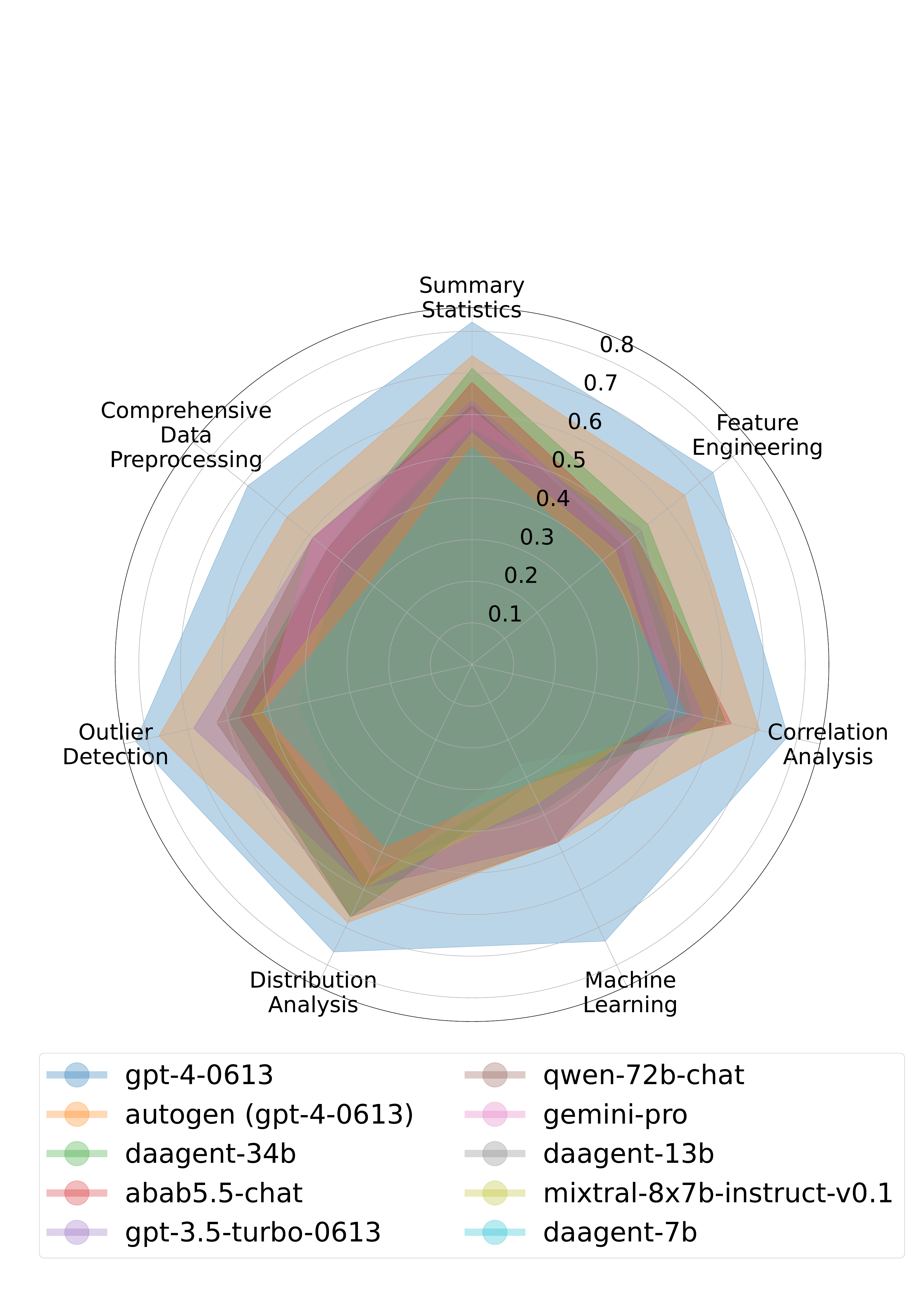}
  \caption{Accuracy of top 10 models in DAEval on the questions involved different concepts.}
  \label{fig:spider}
\end{figure}

Table \ref{tab:main_results} shows the accuracy of different models in InfiAgent-DABench. We also draw a spider chart to illustrate the performance in questions involved different concepts in Figure \ref{fig:spider}. 

\section{Prompts for DAEval Construction}
\label{appendix:dataset construction}
In this section, we list all the prompts used in the process of question generation.


\subsection{Description Generation}
\label{appendix:description generation}
Table \ref{tab:prompts_for_description_generation} shows prompts for description generation.

\subsection{Question Generation}
\label{appendix:question generation}
Table \ref{tab:prompts_for_question_generation} shows prompts for question generation. We prompt GPT-4 in the following way: 1) Three simple questions containing only one concept. 2) Three complex questions involving one or multiple concepts. 3) Three more complex questions involving multiple concepts.

\subsection{Constraints and Format Requirements Generation}
\label{appendix:Constraints and Format requirements Generation}
Table \ref{tab:prompts_for_constrains_generation} shows prompts for constraints and format requirements generation.

\section{Detailed Information about Key Concepts}
\label{appendix:concepts and explanations}
Table \ref{tab:concepts} shows key concepts on data analysis and corresponding explanations obtained from expert interview. We use GPT-3.5 to help us summarize the interview scripts. We could not directly publicize them for some reasons.

\section{Detailed Information about Domains}
\label{appendix:domains}
We cluster the CSV files into 9 domains with the help of GPT-3.5. Table \ref{tab:categories} shows the names of domain and corresponding explanations.

\section{Detailed Information about Pilot Experiments}
\label{appendix:pilot}
We randomly sample 119 questions and corresponding answers from GPT-4, GPT-3.5, DAAgent-34B and ask both human experts and GPT-4 to evaluate based on the same criteria. Each answer is rated based on the level of resolution: 0: Did not solve the problem or barely addressed it, 1: Partly solved the problem, 2: Mostly solved the problem and only have minor errors, 3: Completely and exactly solved the problem. Table \ref{tab:pilot} shows the consistency rate between human experts and GPT-4. We found that when GPT-4 works as an evaluator on data analysis tasks, it only achieves about 67\% consistency with human experts, which is obviously unsatisfactory as an evaluator for a benchmark.

\section{Extended Case Study on Generated Questions}
\label{appendix:case study on generated questions}
Table \ref{tab:question_examples} showcases several questions generated for the file \texttt{test\_ave.csv}. This file meticulously documents the personal information and travel details of passengers on the Titanic, including their survival status, cabin class, name, gender, age, and embarkation point, among other aspects. The generated questions include one or more subquestions according to the specific question and constraints.

\section{Human Assessment}
\label{appendix:human evaluation}

\subsection{Settings for Dataset Quality Assessment}
\label{appendix:questionnaire}
For a high-quality dataset, we invite 3 human experts with math and programming skills to assess the quality of files, questions and labels of all the data samples. For each part, we define one or more metrics. For each metric, to a precise assessment, we decompose it into several yes-or-no questions for experts. Table \ref{tab:questionnaire} shows our questionnaire, including questions and expected response formats. We invite 3 experts with math and programming skills to assess all data samples in DAEval. Only all yes-or-no questions of one metric are positive, the assessment results for this metric is positive. We report the percentage of positiveness on each metric and an overall percentage of positiveness on all the metrics. Only data samples positive on all the metrics are kept. Otherwise, we asked experts to compare difficulty on 216 anonymous question pairs (108 with single concept and 108 with multiple concepts) sampled from human-made and GPT-4 generated questions. Questions with single concept are only compared with other single ones, the same for those with multiple concepts. Table \ref{tab:questionnaire_human_vs_gpt} shows the questionnaire to experts.

\subsection{Bad Case Analysis}
\label{appendix:dataset error analysis}
Table \ref{tab:badcase} presents two bad cases found from human assessment. In the first case, we found all values in ``Type'' column in \texttt{2019-08\_edges.csv} are the same, which means using Random Forest Classifier to predict it here is meaningless. This question doesn't meet our criteria on Alignment. In the second case, the question asks with ``the given dates'' but not provides with the specific dates in the following constraints, which means the question doesn't meet our criteria on Restrictiveness.

\subsection{Settings for Comparison between Human vs GPT-4}
\label{appendix:human_vs_gpt4}
We invite 3 experts with data analysis skills to write 54 questions associated with 9 CSV files. For each file, 3 questions with single concept and 3 with multiple concepts are written. Combined with questions from GPT-4 on the same files, we form a dataset with 102 questions for assessment. Additionally, for difficulty, we randomly sample 216 question pairs on human-made and GPT-4 generated questions. All questions are anonymous to experts in this stage.

\section{Method for Assessing Difficulty} 
\label{appendix:methodology}
For the difficulty assessment of DAEval, we randomly sample 80 groups from the dataset, each comprising 5 questions. GPT-4 ranks the questions within each group according to their difficulty, based on criteria such as data processing complexity, domain-specific knowledge, and coding sophistication. This ranking process is repeated 100 times for each group to ensure the robustness of the results. Each question has a score based on its ranking, with the most challenging question in a group earning 5 points and the least challenging 0 points. An average score was then calculated for each question, leading to a numerical difficulty ranking. These rankings allowed for the classification of questions into three levels: easy (0 to 1.667), medium (1.667 to 3.335), and hard (3.335 to 5).

\section{Statistics on DAInstruct}
\label{appendix:statistics on DAInstruct}
DAInstruct includes 5131 data samples involved 631 CSV files, with the average length of ReAct conversations being 5.18. 

\section{Prompts for DAInstruct Construction}
\label{appendix:prompts_for_DAInstruct_construction}
\subsection{Question Generation}
Table \ref{tab:keywords_and_prompts} lists keywords and prompts used for generating questions.

\subsection{Filtering}
Table \ref{tab:prompts_for_filtering} lists prompts used for filtering.

\subsection{Content Moderation}
Table \ref{tab:prompts_for_content_moderation} lists prompts used for content moderation.

\section{Instruction-tuning data collection}
\label{appendix:react_process}
In our data collection method, GPT-4 engages iteratively in problem-solving with our agent framework. The conversation begins with presenting a CSV file and an associated question. GPT-4 processes this and proposes code solutions, which are then executed. The results of the execution are fed back into the conversation, enabling GPT-4 to refine its responses and propose further iterations. This iterative cycle continues until GPT-4 determines that the accumulated information suffices to conclusively answer the problem. A critical aspect of this process is the design of the prompts, which directs GPT-4's outputs towards effective problem-solving. Table \ref{tab:prompts_for_instruction_tuning_data_collection} lists prompts used in the process. To counter the challenge of hallucination, where GPT-4 might generate fictitious results, explicit instructions are included in the prompt to prevent GPT-4 from executing the code.

\section{Case Study on DAInstruct}
\label{appendix:case study on DAInstruct}
Table \ref{tab:case_study_on_dainstruct} shows a case study on DAInstruct, highlighting a dialogue between GPT-4 and the user, which poses questions and executes generated code. This iterative conversation continues until GPT-4 reaches a conclusive solution.

\section{Implementation Details for Evaluation}
\label{appendix:implementations}
Full prompts for evaluation can be found in Appendix \ref{appendix:prompts for evaluation}. We use accuracy as the metric, which is the proportion of questions for which all sub-questions are answered correctly. We use regular expression matching to draw the answer enclosed in ``answer\_name[answer]'' and the exact match to evaluate the performance. We set temperature 0.2, top\_p 1.0 with nucleus sampling and frequency penalty 0.0 for all the models in the experiments. All instances of GPT-3.5 and GPT-4 used in this paper, unless specifically indicated otherwise, are referred to as \texttt{gpt-3.5-turbo-0613} and \texttt{gpt-4-0613}\footnote{\url{https://platform.openai.com/docs/models}} respectively.

\subsection{Prompts for Evaluation}
\label{appendix:prompts for evaluation}
Table \ref{tab:prompts_for_evaluation} shows standard prompt templates for evaluation. In addition, we slightly modify the standard templates for Abab5.5-Chat to make its output meet the format requirements of the agent framework. Specifically, we add the following prompt: ``Note that you must output the string `I now know the final answer' before the final answer.''.

\section{Details for the Reformatting}
\label{appendix:reformat}
In this section, we conduct ablation studies to demonstrate the necessity of adding a reformatting step after models' original responses, since most models have difficulties in strictly following the format requirements, while it is crucial for a closed-form evaluation. In addition, we conduct manual check to ensure this reformatting step doesn't effect the evaluation results by introducing external errors. We use \texttt{gpt-3.5-turbo-16k-0613} on reformatting. Prompts for reformatting are in Appendix \ref{appendix:prompts_for_reformat}.  

\subsection{Ablation Studies}
We add an additional reformatting step with GPT-3.5 to transform original responses into required formats. We conduct ablation studies in two settings: 1) w/o reformatting: directly prompt models with question, constraints and format requirements at one stage 2) w/ reformatting: first prompt models with question and constraints, and then use GPT-3.5 to reformat the original responses, with GPT-4, Qwen-72B-Chat, Qwen-14B-Chat and Mixtral-7B-Instruct-v0.2. Table \ref{tab:reformat_ablation} shows the results that even the performance of most powerful models like GPT-4 drops about 6\%, and weaker models like Mistral-7B drops up to 32\%, if prompted without reformatting, indicating the necessity of adding a reformatting step for evaluation.

\subsection{Manual Check}
We randomly sample 50 questions with corresponding original responses from models as well as responses processed by reformatting step and manually check if this step could introduce extra errors for evaluation. We found GPT-3.5 can reformat the responses into required formats in all samples, which means this extra step usually will not effect the results.

\subsection{Prompts for reformatting}
\label{appendix:prompts_for_reformat}
Table \ref{tab:prompts_for_reformatting} shows prompts used in reformatting step. 

\begin{table*}
\centering
\footnotesize
\begin{tabular}{lcc}
\toprule
\textbf{Model} & \textbf{\# Params. (in B)} & \textbf{Accuracy (\%)} \\
\midrule
gpt-4-0613 \citep{openai2023gpt4} & / & 78.99 \\
autogen$^\dag$ (gpt-4-0613) \citep{wu2023autogen} & / & 71.48 \\
\textbf{daagent-34b} & 34 & 64.59 \\
abab5.5-chat \citep{minimax2023abab5.5} & / & 61.48 \\
gpt-3.5-turbo-0613 \citep{openai2023gpt3.5} & / & 60.70 \\
qwen-72b-chat \citep{bai2023qwen} & 72 & 59.92 \\
gemini-pro \citep{team2023gemini} & / & 56.42 \\
\textbf{daagent-13b} & 13 & 55.86 \\
qwen-agent$^\dag$ (qwen-72b-chat) \citep{bai2023qwen} & 72 & 55.25 \\
mixtral-8x7b-instruct-v0.1$^{*}$ \citep{mistral2023} & 46.7 (12.9) & 52.92 \\
\textbf{daagent-7b} & 7 & 51.36 \\
xagent$^\dag$ (gpt-4-0613) \citep{xagent2023} & / & 47.52 \\
claude-2.1 \citep{claude2023} & / & 47.47 \\
deepseek-coder-33b-instruct \citep{deepseek-coder} & 33 & 46.09 \\
phind-codellama-34b-v2 \citep{phind2023phindcodellama} & 34 & 43.87 \\
xwincoder-34b \citep{xwin-lm} & 34 & 41.25 \\
mistral-7b-instruct-v0.2 \citep{mistral2023} & 7 & 38.67 \\
yi-34b \citep{2023yi} & 34 & 38.34 \\
qwen-14b-chat \citep{bai2023qwen} & 14 & 37.50 \\
vicuna-13b \citep{vicuna2023} & 13 & 35.83 \\
agentlm-70b \citep{zeng2023agenttuning} & 70 & 33.46 \\
codellama-34b-instruct \citep{rozière2023code} & 34 & 31.13 \\
vicuna-13b-v1.5 \citep{vicuna2023} & 13 & 27.31 \\
qwen-7b-chat \citep{bai2023qwen} & 7 & 27.27 \\
vicuna-7b \citep{vicuna2023} & 7 & 26.95 \\
codellama-13b-instruct \citep{rozière2023code} & 13 & 26.67 \\
codellama-7b-instruct \citep{rozière2023code} & 7 & 24.61 \\
internlm-chat-20b \citep{2023internlm} & 20 & 24.51 \\
agentlm-13b \citep{zeng2023agenttuning} & 13 & 24.31 \\
wizard-coder-13b \citep{luo2023wizardcoder} & 13 & 23.14 \\
wizard-coder-7b \citep{luo2023wizardcoder} & 7 & 22.83 \\
xwincoder-13b \citep{xwin-lm} & 13 & 19.69 \\
agentlm-7b \citep{zeng2023agenttuning} & 7 & 17.46 \\
chatglm3-6b \citep{zeng2022glm} & 6 & 18.10 \\
xwincoder-7b \citep{xwin-lm} & 7 & 10.71 \\
baichuan2-13b-chat \citep{yang2023baichuan} & 13 & 9.34 \\
baichuan2-7b-chat \citep{yang2023baichuan} & 7 & 8.95 \\
\bottomrule
\end{tabular}
\caption{\textbf{Leaderboard of InfiAgent-DABench.} Our DAAgent models are marked bold. $^{*}$Mixtral is a model utilizing sparse mixture-of-experts (SMoE) architecture, with 46.7B total parameters but only uses 12.9B parameters per token. $^\dag$All models except ready-made agent frameworks are equipped with our framework to interacter with files and code interpreter. We present LLMs those ready-made frameworks based in the experiments in parentheses.
} 
\label{tab:main_results}
\end{table*}

\begin{table*}
\centering
\begin{tabular}{p{\textwidth}}
\toprule
\textbf{Prompts for Description Generation} \\
\midrule
Here's the example of a detailed description of the table's header: ``\\
The dataset contains 4599 entries and 5 columns. Here is a detailed description of each column: \\
1. Crime ID \\
Unique identifier for each crime. \\ 
Contains null values. \\
Data type: Object (String) \\
2. Month \\
The month when the crime was reported. \\
No missing values. \\
Data type: Object (String, in the format YYYY-MM) \\
3. Reported by \\
The police force that the crime was reported to. \\
No missing values. \\
Data type: Object (String) \\
4. Falls within \\
The higher-level police force that the crime report falls within.
No missing values. \\
Data type: Object (String) \\
5. Longitude \\
The longitude where the crime occurred. \\
Contains null values. \\
Data type: Float64''. \\
This is the information of one table:\\
\{info\}\\
\{head\}.\\ 
Please provide a detailed description of this table's header.\\
\bottomrule
\end{tabular}
\caption{Prompts for description generation.}
\label{tab:prompts_for_description_generation}
\end{table*}
\begin{table*}
\centering
\begin{tabular}{p{\textwidth}}
\toprule
\textbf{Prompts for Question Generation} \\
\midrule
You are now a teacher responsible for creating data analysis questions. Please provide 3 possible question prompts based on the given scenario information. Ensure that the answers to these questions only require information within the given scenario and need to be solved using Python code. In data analysis, there are several key concepts: 
\{key\_concepts\}. Please provide questions that are \{complexity\}. You do not need to answer the questions themselves. Scenario: You have a file \{file\_name\}. Here is a detailed description of each column (head) in the provided table: \{file\_description\}. Please use \textbackslash Question\{question\} to wrap each of the 3 questions separately. Please use \textbackslash Concepts\{concepts\} to point out which concepts this question wants to test. \\
\bottomrule
\end{tabular}
\caption{Prompts for question generation.}
\label{tab:prompts_for_question_generation}
\end{table*}
\begin{table*}
\centering
\begin{tabular}{p{\textwidth}}
\toprule
\textbf{Prompts for Constrains and Format Generation} \\
\midrule

Please keep questions unchanged but \\1) Add more constraints to the question to make sure the question has single fixed answer. Specify the requirements completely and precisely, so that the students don't need to do any assumption or determine the statistical methods or parameters by themselves. Please use \textbackslash Constraints\{constraints\} to wrap all the constraints. \\2) Specify the desired output format requirements for the questions to make sure the answer can be easily checked. For a single answer, use format like @answer\_name[answer] for an easy parsing. You can define the "answer\_name" according to the specific question. "answer" should have a given format and a clear domain of value. Please use \textbackslash Format\{format\} to wrap all the output requirements mentioned above. Some good demonstrations for constraints and format are: 
```\\
\textbackslash Constraints\{\{
Calculate the Pearson correlation coefficient (r) to assess the strength and direction of the linear relationship between danceability and energy.
Assess the significance of the correlation using a two-tailed test with a significance level (alpha) of 0.05.
Report the p-value associated with the correlation test.
Consider the relationship to be linear if the p-value is less than 0.05 and the absolute value of r is greater than or equal to 0.5.
Consider the relationship to be nonlinear if the p-value is less than 0.05 and the absolute value of r is less than 0.5.
If the p-value is greater than or equal to 0.05, report that there is no significant correlation.
\}\}\\
\textbackslash Format\{\{
@correlation\_coefficient[r\_value]
@p\_value[p\_value]
@relationship\_type[relationship\_type]
where "r\_value" is a number between -1 and 1, rounded to two decimal places.
where "p\_value" is a number between 0 and 1, rounded to four decimal places.
where "relationship\_type" is a string that can either be "linear", "nonlinear", or "none" based on the conditions specified in the constraints.
\}\}
```. \\The constraints and formats can be adjusted according to the specific problem to ensure diversity, but strictly use correct symbol as the demonstration like @answer\_name[answer].
The answer should be: \textbackslash Question\{question\}
\textbackslash Concepts\{concepts\}
\textbackslash Constraints\{constraints\}
\textbackslash Format\{format\}.\\
\bottomrule
\end{tabular}
\caption{Prompts for constrains and format requirements generation.}
\label{tab:prompts_for_constrains_generation}
\end{table*}

\begin{table*}
\centering
\begin{tabular}{p{\textwidth}}
\toprule
\textbf{Concepts and corresponding explanations} \\
\midrule
\textbf{Concept}: Summary Statistics \\
\textbf{Explanation}: To start, calculate key summary statistics such as mean, median, standard deviation, minimum, maximum, etc., to obtain an overview of the data. These statistics help understand the central tendencies and distribution range of the data. \\
\midrule
\textbf{Concept}: Distribution Analysis \\
\textbf{Explanation}: Explore the distribution of the data, including checking if it adheres to a normal distribution, exhibits skewness, heavy tails, or bimodality. This helps in selecting appropriate statistical methods and models. \\
\midrule
\textbf{Concept}: Correlation Analysis \\
\textbf{Explanation}: Analyze the correlations between various variables. This assists in determining linear or nonlinear relationships between variables. \\
\midrule
\textbf{Concept}: Outlier Detection \\
\textbf{Explanation}: Identify and deal with outliers, as they can potentially disrupt data analysis and modeling. Methods such as box plots, Z-scores, or specialized outlier detection algorithms can be employed for outlier identification. \\
\midrule
\textbf{Concept}: Comprehensive Data Preprocessing \\
\textbf{Explanation}: Comprehensive data preprocessing is a fundamental step in the data analysis workflow, encompassing data cleaning, transformation, and the handling of missing values. It begins with data cleaning, a process focused on ensuring the accuracy and consistency of the data by identifying and rectifying errors, duplications, and inconsistencies. In tandem, data transformation adjusts the data’s format and structure, which includes normalization, encoding categorical variables, and generating derived features that better represent the underlying phenomena for analysis. Integral to this preprocessing stage is the management of missing values, which may involve strategies such as deletion, imputation, or interpolation, depending on the nature and extent of the missing data. \\
\midrule
\textbf{Concept}: Feature Engineering \\
\textbf{Explanation}: New features can be generated or existing ones transformed to extract more information or improve model performance. \\
\midrule
\textbf{Concept}: Machine Learning \\
\textbf{Explanation}: Harness algorithms to classify data into categories, make predictions through regression, discover hidden patterns using clustering techniques, and even uncover insights from time series data. Explore the fundamentals of model training, evaluation, and practical applications, enabling to extract valuable information and make data-driven decisions across a wide range of analytical tasks. \\
\bottomrule
\end{tabular}
\caption{\label{tab:concepts}
Concepts and corresponding explanations concluded from the expert interview.
}
\end{table*}
\begin{table*}
\centering
\begin{tabular}{p{0.33\textwidth}p{0.67\textwidth}}
\toprule
\textbf{Category} & \textbf{Description} \\
\midrule
Finance and Economics & Covers financial market trends, economic indicators, real estate prices, and investment analysis, essential for understanding economic conditions and market dynamics. \\
\midrule
Health and Medical & Includes health insurance, disease statistics, hospital performance, and pharmaceutical research, key to evaluating public health policies and medical services effectiveness. \\
\midrule
Demographics and Social Science & Comprises education levels, employment status, demographics, and social welfare, fundamental for analyzing social structures and social science issues. \\
\midrule
Marketing and Consumer Behavior & Focuses on market segmentation, consumer preferences, and purchasing behavior, providing support for market strategy formulation and consumer behavior research. \\
\midrule
Energy and Environmental Monitoring & Contains energy consumption, environmental monitoring, and meteorological information, crucial for resource management, environmental policy, and climate change research. \\
\midrule
Transportation, Logistics, and Tourism & Involves traffic management, logistics efficiency, and tourism statistics, key to improving the efficiency of transportation and logistics as well as the management of the tourism industry. \\
\midrule
Culture, Entertainment, and Media & Includes music, gaming, media, and cultural activities, used to analyze trends in the entertainment and cultural industries. \\
\midrule
Scientific Research and Technology & Covers a broad range of scientific research and technological development, from genetics to engineering, suitable for scientific discovery and technological advancement. \\
\midrule
Other Categories & Includes data analysis, machine learning, statistical models, specialized studies, and datasets that do not fall under any of the above categories, applicable to a variety of interdisciplinary data analysis and research work. \\
\bottomrule
\end{tabular}
\caption{Domains and corresponding descriptions.}
\label{tab:categories}
\end{table*}
\begin{table*}
\centering
\begin{tabular}{lllll}
\toprule
\textbf{Response Model} & GPT-4 & GPT-3.5 & DAAgent-34B & Avg.\\
\midrule
\textbf{Consistency (\%)} & 66.4 & 68.9 & 65.5 & 66.9 \\
\bottomrule
\end{tabular}
\caption{\label{tab:pilot}
\textbf{Consistency rate between human experts and GPT-4 on the evaluation of open-ended questions.} We gather responses from GPT-4, GPT-3.5 and DAAgent-34B on solving open-ended data analysis problems. We ask both human experts and GPT-4 to rate responses based on the same criteria and calculate the percentage of rating both sides reach consensus on.
}
\end{table*}
\begin{table*}
\centering
\begin{tabular}{p{\textwidth}}
\toprule
\textbf{Generated questions for test\_ave.csv} \\
\midrule
\textbf{Question}: Calculate the mean fare paid by the passengers. \\
\textbf{Concepts}: Summary Statistics \\
\textbf{Constraints}: Calculate the mean fare using Python's built-in statistics module or appropriate statistical method in pandas. Rounding off the answer to two decimal places. \\
\textbf{Format}: \textcolor{red}{@mean\_fare[mean\_fare\_value]} where "mean\_fare\_value" is a floating-point number rounded to two decimal places. \\
\textbf{Labels}: \textcolor{red}{@mean\_fare[34.65]} \\
\midrule
\textbf{Question}: Apply the linear regression algorithm from the sklearn library to predict whether a passenger survived or not based on the features 'Pclass', 'Sex', 'Age', 'SibSp', 'Parch', 'Fare', and 'Embarked'. Encode 'Sex' and 'Embarked' to numerical values before applying the model. Split the dataset into a training set (80\%) and a testing set (20\%), train the model on the training set, and evaluate its performance on the testing set using the accuracy score. Ensure that the train\_test\_split function's random\_state parameter is set to 42 for consistency. \\
\textbf{Concepts}: Machine Learning \\
\textbf{Constraints}: Use one-hot encoding for the 'Sex' and 'Embarked' features. Use the "linear regression" model provided by the sklearn library in Python. \\
\textbf{Format}: \textcolor{red}{@prediction\_accuracy[accuracy]}, where "accuracy" is a float number rounded to 2 decimal places and has a range of 0.0 to 1.0. \\
\textbf{Labels}: \textcolor{red}{@prediction\_accuracy[0.78]} \\
\midrule
\textbf{Question}: Generate a new feature called "FamilySize" by summing the "SibSp" and "Parch" columns. Then, calculate the Pearson correlation coefficient (r) between the "FamilySize" and "Fare" columns. \\
\textbf{Concepts}: Feature Engineering, Correlation Analysis \\
\textbf{Constraints}: Create a new column 'FamilySize' that is the sum of 'SibSp' and 'Parch' for each row.
Calculate the Pearson correlation coefficient between 'FamilySize' and 'Fare'
Do not perform any further data cleaning or preprocessing steps before calculating the correlation. \\
\textbf{Format}: \textcolor{red}{@correlation\_coefficient[r\_value]}
where "r\_value" is the Pearson correlation coefficient between 'FamilySize' and 'Fare', a number between -1 and 1, rounded to two decimal places. \\
\textbf{Labels}: \textcolor{red}{@correlation\_coefficient[0.21]} \\
\midrule
\textbf{Question}: Create a new column called "AgeGroup" that categorizes the passengers into four age groups: 'Child' (0-12 years old), 'Teenager' (13-19 years old), 'Adult' (20-59 years old), and 'Elderly' (60 years old and above). Then, calculate the mean fare for each age group. \\
\textbf{Concepts}: Feature Engineering, Summary Statistics \\
\textbf{Constraints}: Make sure to round the mean fare of each group to 2 decimal places. \\
\textbf{Format}: \textcolor{red}{@mean\_fare\_child[mean\_fare]}, \textcolor{red}{@mean\_fare\_teenager[mean\_fare]}, \textcolor{red}{@mean\_fare\_adult[mean\_fare]}, \textcolor{red}{@mean\_fare\_elderly[mean\_fare]}, where "mean\_fare" is a float number rounded to 2 decimal places. \\
\textbf{Labels}: \textcolor{red}{@mean\_fare\_elderly[43.47]}, \textcolor{red}{@mean\_fare\_teenager[31.98]}, \textcolor{red}{@mean\_fare\_child[31.09]}, \textcolor{red}{@mean\_fare\_adult[35.17]} \\
\midrule
\textbf{Question}: Perform a distribution analysis on the 'Fare' column for each passenger class ('Pclass') separately. Calculate the mean, median, and standard deviation of the fare for each class. Interpret the results in terms of the different passenger classes. \\
\textbf{Concepts}: Distribution Analysis, Summary Statistics \\
\textbf{Constraints}: Keep all numerical values rounded to 2 decimal points. The population standard deviation should be calculated. \\
\textbf{Format}: \textcolor{red}{@mean\_fare\_class1[mean\_fare]}, \textcolor{red}{@median\_fare\_class1[median\_fare]}, \textcolor{red}{@std\_dev\_fare\_class1[std\_dev]}, \textcolor{red}{@mean\_fare\_class2[mean\_fare]}, \textcolor{red}{@median\_fare\_class2[median\_fare]}, \textcolor{red}{@std\_dev\_fare\_class2[std\_dev]}, \textcolor{red}{@mean\_fare\_class3[mean\_fare]}, \textcolor{red}{@median\_fare\_class3[median\_fare]}, \textcolor{red}{@std\_dev\_fare\_class3[std\_dev]}, where "mean\_fare", "median\_fare", and "std\_dev" are statistical measures in float format rounded to 2 decimal places. \\
\textbf{Labels}: \textcolor{red}{@median\_fare\_class1[69.30]}, \textcolor{red}{@median\_fare\_class2[15.05]}, \textcolor{red}{@std\_dev\_fare\_class1[80.86]}, \textcolor{red}{@mean\_fare\_class3[13.23]}, \textcolor{red}{@std\_dev\_fare\_class2[13.19]}, \textcolor{red}{@mean\_fare\_class2[21.47]}, \textcolor{red}{@std\_dev\_fare\_class3[10.04]}, \textcolor{red}{@mean\_fare\_class1[87.96]}
\\
\bottomrule
\end{tabular}
\caption{\label{tab:question_examples}
\textbf{Extended case study on generated questions.} All ``@answer\_name[answer]'' in \textbf{Format} and \textbf{Labels} are marked in \textcolor{red}{red}.
}
\end{table*}
\begin{table*}[ht]
\centering
\begin{tabular}{p{0.2\textwidth} p{0.8\textwidth}} \\
\toprule
\textbf{Category} & \textbf{Question and Answer} \\
\midrule
File & 
\textbf{1.1} (\textcolor{red}{Suitableness}): Does the csv file represent a meaningful scenario for data analysis? (Yes/No, reasons) \\
& \textbf{1.2}: Is the file's \textit{Category} correct? (Yes/No, correct category) \\
& \textbf{1.3}: Does the file involve sensitive attributes or offensive content? (No/Yes, keywords) \\
\midrule
Question & 
\textbf{2.1} (\textcolor{red}{Reasonableness}): Can \textit{Question} express a data analysis problem naturally and reasonably? (Yes/No, reasons) \\
& \textbf{2.2} (\textcolor{red}{Reasonableness}): Are \textit{Constraints} limitations on \textit{Question}? (Yes/No, reasons) \\
& \textbf{2.3} (\textcolor{red}{Reasonableness}): Do \textit{Constraints} match \textit{Question}? (Yes/No, reasons) \\
& \textbf{2.4} (\textcolor{red}{Reasonableness}): Does \textit{Format} specify value type and range for all “@answer\_name[answer]”? (Yes/No, non-compliant items) \\
& \textbf{2.5} (\textcolor{red}{Reasonableness}): Does \textit{Format} match \textit{Question}/\textit{Constraints}? (Yes/No, reasons) \\
& \textbf{2.6}: Do \textit{Question}/\textit{Constraints} involve sensitive content? (No/Yes, keywords) \\
& \textbf{2.7} (\textcolor{red}{Value}): Do \textit{Question}/\textit{Constraints} have value for data analysis? (Yes/No, reasons) \\
& \textbf{2.8}: Do \textit{Question}/\textit{Constraints}/\textit{Format} match \textit{Concepts}? (No/Yes, necessary concepts) \\
& \textbf{2.9} (\textcolor{red}{Alignment}): Does the table have required data columns for \textit{Question}/\textit{Constraints}? (Yes/No, missing columns) \\
& \textbf{2.10} (\textcolor{red}{Alignment}): Do data types and ranges in columns match \textit{Question}/\textit{Constraints}/\textit{Format} requirements? (Yes/No, non-compliant columns) \\
\midrule
Label & 
\textbf{3.1} (\textcolor{red}{Correctness}): Is \textit{Label} correct? (Yes/No, incorrect “@answer\_name[name]”) \\
& \textbf{3.2} (\textcolor{red}{Correctness}): Does \textit{Label} comply with \textit{Format} requirements? (Yes/No, non-compliant “@answer\_name[name]”) \\
& \textbf{3.3} (\textcolor{red}{Restrictiveness}): Are \textit{Question}/\textit{Constraints}/\textit{Format} sufficient for unique correct answers in \textit{Label}? (Yes/No, reasons) \\
\bottomrule
\end{tabular}
\caption{\textbf{Questionnaire to experts for assessing data quality on DAEval.} Metrics (\textcolor{red}{Suitableness}, \textcolor{red}{Reasonableness}, \textcolor{red}{Value}, \textcolor{red}{Alignment}, \textcolor{red}{Correctness}, \textcolor{red}{Restrictiveness}) are marked in red, and dataset components (\textit{Question}, \textit{Format}, \textit{Constraints}, \textit{Concepts}, \textit{Label}) are in italics. Human experts should give answers according to requirements in parentheses.}
\label{tab:questionnaire}
\end{table*}
\begin{table*}[ht]
\centering
\begin{tabular}{p{\textwidth}} \\
\toprule
\textbf{Question and Answer} \\
\midrule
\textbf{1} (\textcolor{red}{Reasonableness}): Is the expression natural and reasonable, and is it a question around data analysis? (5 points: Very natural and reasonable, 1 point: Very unnatural and unreasonable) \\
\textbf{2} (\textcolor{red}{Value}): Is it a valuable question? (5 points: Very valuable, 1 point: Not valuable at all) \\
\textbf{3} (\textcolor{red}{Restrictiveness}): Can it strictly limit the answer to the question as unique? (5 points: Completely strict, 1 point: Not strict at all) \\
\textbf{4} (\textcolor{red}{Alignment}): Can it align with the data? Can the data needed to solve the question be found in the table, and does it meet the requirements? (5 points: Completely aligned, 1 point: Not aligned at all) \\
\textbf{5} (\textcolor{red}{Difficulty}): Among two questions (A and B), which one is more difficult? (A/B/Close) \\
\bottomrule
\end{tabular}
\caption{\textbf{Questionnaire to experts for comparing human-made and GPT-4 generated questions.} Metrics (\textcolor{red}{Reasonableness}, \textcolor{red}{Value}, \textcolor{red}{Restrictiveness}, \textcolor{red}{Alignment}, \textcolor{red}{Difficulty}) are marked in red. Assessors should rate each question according to requirements in parentheses.}
\label{tab:questionnaire_human_vs_gpt}
\end{table*}

\begin{table*}
\centering
\begin{tabular}{p{\textwidth}}
\toprule
\textbf{Bad Cases} \\
\midrule
\textbf{File}: 2019-08\_edges.csv \\
\textbf{Question}: Build a machine learning model to \textcolor{red}{predict the "Type"} based on the other columns in the dataset. \\
\textbf{Constraints}: Split the dataset into 70\% training and 30\% testing. Use Random Forest Classifier for building the model. Use "Source", "Target", "Weight", "lng\_org", "lat\_org", "lng\_dest", "lat\_dest", and the newly created "Distance" as features for training the model. Evaluate the model's performance using accuracy score, precision, recall, and F1 score. \\
\textbf{Format}: @accuracy\_score[accuracy], @precision[precision], @recall[recall], @f1\_score[f1] where "accuracy", "precision", "recall", and "f1" are the respective evaluation metrics for the model rounded to four decimal places. \\
\textbf{Label}: @precision[1.0000], @recall[1.0000], @accuracy\_score[1.0000], @f1\_score[1.0000] \\
\midrule
\textbf{File}: tr\_eikon\_eod\_data.csv \\
\textbf{Question}: Explore the distribution of the CBOE Volatility Index (.VIX) on \textcolor{red}{the given dates}. \\
\textbf{Constraints}: Use the numpy histogram function to explore the distribution. Set the bin separator array to [10, 20, 30, 40, 50, 60, 70]. Only consider non-null values. \\
\textbf{Format}: @VIX\_bin\_counts[bin\_counts] where "bin\_counts" is an array of 6 integers representing the count of .VIX values in each bin. \\
\textbf{Label}: @VIX\_bin\_counts[1623, 352, 89, 15, 0, 0]\\
\bottomrule
\end{tabular}
\caption{\textbf{Bad cases found from human assessment.} The questionable parts of the cases are marked in \textcolor{red}{red}.}
\label{tab:badcase}
\end{table*}

\begin{table*}[ht]
\centering
\begin{tabular}{p{0.2\textwidth}p{0.8\textwidth}}
\toprule
\textbf{Category} & \textbf{Description} \\
\midrule
Domain Keywords & Data Cleaning \& Preprocessing, Data Exploration, Text Data Analysis, Queries, Data Visualization Recommendations, Ethics and Bias, Documentation \& Communication, Data Integrity Analysis, Data Pattern Identification \\
\midrule
Question Prompt & \{csv\_summary\} \newline Please come up with a data analysis question preferably in \{domain keyword\} (if no proper question you can think of, propose a general question) that can use python code to solve. The question is preferred to have a close-formed solution. \newline Respond with the question ONLY. \\
\bottomrule
\end{tabular}
\caption{List of keywords and the prompt to generate questions.}
\label{tab:keywords_and_prompts}
\end{table*}

\begin{table*}[ht]
\centering
\begin{tabular}{p{0.2\textwidth}p{0.8\textwidth}}
\toprule
\textbf{Filtering Aspect} & \textbf{Prompt} \\
\midrule
Python Code Relevance & \{conversation\} Please read the dialogue carefully and judge whether the provided Python code is syntactically correct and relevant to the task. \\
\midrule
System Notifications & \{conversation\} Assess if the dialogue accurately interprets system notifications and outputs in the Python sandbox environment. \\
\midrule
Conciseness & \{conversation\} Evaluate if the response is concise and focused on the task objectives using the Python sandbox. \\
\midrule
Sandbox Environment & \{conversation\} Verify if the dialogue indicates the use of a real Python sandbox environment. \\
\midrule
Solution Provision & \{conversation\} Determine if the dialogue provides a clear solution to the problem using the Python sandbox environment. \\
\bottomrule
\end{tabular}
\caption{Prompts used for GPT-4 filtering with enhanced criteria.}
\label{tab:prompts_for_filtering}
\end{table*}

\begin{table*}[ht]
\centering
\begin{tabular}{p{0.2\textwidth}p{0.8\textwidth}}
\toprule
\textbf{Moderation Aspect} & \textbf{Prompt} \\
\midrule
Checking the CSV Data & \{csv.head().to\_string()\} Do you think the above csv file contains any sensitive information including but not limited to Drug, Marijuana, Weeds, Cannabis, Violence, Sexual assault, Politics? Yes or no? If yes, what does it contain? \\
\midrule
Checking the Conversation & \{conversation\} Do you think the above dialogue contains any sensitive information including but not limited to Drug, Marijuana, Weeds, Cannabis, Violence, Sexual assault, Politics? Yes or no? If yes, what does it contain? \\
\bottomrule
\end{tabular}
\caption{Prompts used for comprehensive content moderation.}
\label{tab:prompts_for_content_moderation}
\end{table*}
\begin{table*}[ht]
\centering
\begin{tabular}{p{0.2\textwidth}p{0.8\textwidth}}
\toprule
\textbf{Steps} & \textbf{Prompt} \\
\midrule
Initial Prompt & Answer the following questions as best you can. You have access to the following tools:\newline python\_code\_sandbox[input]: this tool can help to run python script with python code as input.\newline Use the following format:\newline Question: the input question you must answer\newline Thought: you should always think about what to do\newline Action: the action to take, should be one of [python\_code\_sandbox]\newline Action Input: the input to the action\newline Observation: the result of the action\newline ... (this Thought/Action/Action Input/Observation can repeat N times)\newline \#Thought: I now know the final answer\newline Final Answer: the final answer to the original input question\newline For all outputs in code, THE print() function MUST be called\newline For example, when you need to call df.head(), please use print(df.head())\newline If you have any files outputted write them to "./output/"\newline Be mindful that when creating a piece of code for execution in python\_code\_sandbox, DO NOT try to predict or complete its output.\newline Cease generating further and await the execution outcome from python\_code\_sandbox.\newline Please note that any text beginning with "[SYSTEM NOTIFICATION] Running the above tool ..." is an output from the python\_code\_sandbox execution. You should NEVER try to generate it.\newline Begin!\newline Question: User uploaded the following files: \{filename\} \newline \{question\} \newline Thought: \\
\midrule
Reply Prompt & Observation:\newline [SYSTEM NOTIFICATION] Running the above tool with the following response:  \{STDOUT\} \newline Thought: \\
\bottomrule
\end{tabular}
\caption{Prompts for data collection.}
\label{tab:prompts_for_instruction_tuning_data_collection}
\end{table*}

\begin{table*}[ht]
\centering
\begin{tabular}{p{0.1\textwidth} p{0.9\textwidth}}
\toprule
\textbf{Speaker} & \textbf{Conversation} \\
\midrule
User & Answer the following questions as best you can. You have access to the following tools: \newline
python\_code\_sandbox[input]: this tool can help to run python script with python code as input. \newline
... \newline
Question: User uploaded the following files: ['/mnt/ep\_7.csv'] \newline
Compute the average number of lines per scene in the 'ep\_7.csv' dataset. \newline
Thought: \\
\midrule
GPT-4 & To answer this question, I will need to \newline
1. Load the CSV into a pandas DataFrame \newline
2. Analyze the dataset to see what it looks like and identify which column contains the scene information \newline
3. Count how many lines belong to each scene \newline
4. Compute the average number of lines per scene \newline
... \newline
Action: python\_code\_sandbox \newline
Action Input: \newline
\texttt{import pandas as pd \newline
filename = "/mnt/ep\_7.csv" \newline
dataframe = pd.read\_csv(filename) \newline
print(dataframe.head())} \newline
Let's load the file and see what the data looks like. \\
\midrule
User & Observation: \newline
[SYSTEM NOTIFICATION] Running the above tool with the following response: \newline
STDOUT: \newline
[Output of dataframe.head()] \newline
Thought: \\
\midrule
GPT-4 & We can see that the dataset is a DataFrame with columns 'Season', 'Episode', 'Scene', 'Character', and 'Line'. The lines of script are contained in the 'Line' column, and each line is associated with a scene number in the 'Scene' column. Now I can calculate the total number of lines for each scene. \newline
Action: python\_code\_sandbox \newline
Action Input: \newline
\texttt{scene\_lines\_counts = dataframe['Scene'].value\_counts().values \newline
avg\_lines\_per\_scene = scene\_lines\_counts.mean() \newline
print(avg\_lines\_per\_scene)} \newline
Here, I'm counting the number of lines in each scene using the value\_counts function on the 'Scene' column of the dataset. This will return a Series with the scene numbers as the index and the counts as the values. The values() function then gets the counts as a numpy array. The mean() function is then used to compute the average number of lines per scene by averaging the counts.
             \\
\midrule
User & Observation: \newline
[SYSTEM NOTIFICATION] Running the above tool with the following response: \newline
STDOUT: \newline
6.018867924528302 \newline
Thought: \\
\midrule
GPT-4 & I now know the final answer \newline
Final Answer: The average number of lines per scene in the 'ep\_7.csv' dataset is approximately 6.02. \\
\bottomrule
\end{tabular}
\caption{Case study on DAInstruct.}
\label{tab:case_study_on_dainstruct}
\end{table*}

\begin{table*}[ht]
\centering
\begin{tabular}{p{\textwidth}}
\toprule
\textbf{Prompts for Evaluation} \\
\midrule
Answer the following questions as best you can. You have access to the following tools: \{tool\_description\}. Use the following format:
Question: the input question you must answer \\
Thought: you should always think about what to do \\
Action: the action to take, should be one of [\{tool\_names\}] \\
Action Input: the input to the action \\
Observation: the result of the action \\
... (this Thought/Action/Action Input/Observation can repeat N times) \\
\#Thought: I now know the final answer \\
Final Answer: the final answer to the original input question \\
For all outputs in code, THE print() function MUST be called \\
For example, when you need to call df.head(), please use print(df.head()) \\
If you have any files outputted write them to "./output/" \\
Do not use things like plot.show() as it will not work instead write them out "./output/" \\
Be mindful that when creating a piece of code for execution in \{tool\_names\}, \\
DO NOT try to predict or complete its output. \\
Cease generating further and await the execution outcome from \{tool\_names\}. \\
Please note that any text beginning with  \\
"[SYSTEM NOTIFICATION] Running the above tool ..." is an output from the \{tool\_names\} execution. \\
You should NEVER try to generate it. \\
Here is an example of using the \{tool\_names\}:  \\
Action: python\_code\_sandbox \\
Action Input: \\
\texttt{print('hello world')} \\
Now Begin! \\
Question: \{instruction\} \\
Thought: \\
\bottomrule
\end{tabular}
\caption{Prompts for evaluation.}
\label{tab:prompts_for_evaluation}
\end{table*}

\begin{table*}
\centering
\begin{tabular}{lllll}
\toprule
\textbf{Model} & \textbf{w/o reformatting} & \textbf{w/ reformatting} \\
\midrule
gpt-4-0613 & 72.76 & 78.99 \\
qwen-72b-chat & 44.75 & 59.92 \\
qwen-14b-chat & 18.29 & 37.50 \\
mistral-7b-instruct-v0.2 & 6.23 & 38.67 \\
\bottomrule
\end{tabular}
\caption{\textbf{The ablation studies of models with and without reformatting.} We use GPT-4 as an evaluator and conduct performance comparison between several LLMs (also including GPT-4, which means GPT-4 evaluates itself in this setting).}
\label{tab:reformat_ablation}
\end{table*}

\begin{table*}
\centering
\begin{tabular}{p{\textwidth}}
\toprule
\textbf{Prompts for Reformatting} \\
\midrule
You should strictly follow the output requirements in the Format part. Here\'re some examples: \textbackslash Format\{\\
@shapiro\_wilk\_statistic[test\_statistic] \\
@shapiro\_wilk\_p\_value[p\_value]\\
where "test\_statistic" is a number between 0 and 1 representing the Shapiro-Wilk test statistic. Rounding off the answer to two decimal places.\\
where "p\_value" is a number between 0 and 1 representing the p-value from the Shapiro-Wilk test. Rounding off the answer to four decimal places.\}\\
\textbackslash Answer\{\\
@shapiro\_wilk\_statistic[0.56]\\
@shapiro\_wilk\_p\_value[0.0002]\}\\
\textbackslash Format\{\\
@total\_votes\_outliers\_num[outlier\_num]\\
where "outlier\_num" is an integer representing the number of values considered outliers in the \'total\_votes\' column.\}\\
\textbackslash Answer\{\\
@total\_votes\_outliers[10]\}.\\
Your answer should contain all the "@answer\_name[answer]" in the order mentioned, each "answer" should be in the range of value as required.\\
The format requirements of this question is:\\
@median\_tmax["median\_TMAX"]\\
@median\_tmin["median\_TMIN"]\\
@range\_tmax["range\_TMAX"]\\
@range\_tmin["range\_TMIN"]\\
where "median\_TMAX" and "median\_TMIN" are the median temperatures for TMAX and TMIN observation types, respectively. They are numbers between -100 and 100, rounded to two decimal places.\\
where "range\_TMAX" and "range\_TMIN" are the range of temperatures for TMAX and TMIN observation types, respectively. They are numbers greater than 0. Please give your answer: \\
\bottomrule
\end{tabular}
\caption{Prompts for reformatting.}
\label{tab:prompts_for_reformatting}
\end{table*}

\end{document}